RESEARCH PAPER

# Classical Machine Learning: Seventy Years of Algorithmic Learning Evolution

Absalom E. Ezugwu[1*], Yuh-Shan Ho[2*], Ojonukpe S. Egwuche[1], Olufisayo S. Ekundayo[1], Annette Van Der Merwe[3], Apu K. Saha[4], Jayanta Pal[5]

[1] Unit for Data Science and Computing, North-West University, 11 Hoffman Street, Potchefstroom 2520, South Africa

[2] Trend Research Centre, Asia University, No. 500, Lioufeng Road, Taichung 41354, Taiwan

[3] School of Computer Science and Information Systems, North-West University, 11 Hoffman Street, Potchefstroom 2520, South Africa

[4] Department of Mathematics, National Institute of Technology Agartala, Agartala, Tripura, 799046, India

[5] Department of IT, Tripura University, Suryamaninagar, Tripura 799022, India

*Corresponding author: Absalom E. Ezugwu (Email: absalom.ezugwu@nwu.ac.za ; ORCID: 0000-0002-3721-3400)；Yuh-Shan Ho (Email: ysho@asia.edu.tw; ORCID: 0000-0002-2557-8736)

Submitted: May 19, 2023; Revised: Jun 17, 2024; Accepted: Jul 15, 2024

Accepted for publication in: Data Intelligence | MIT Press

**Abstract**. Machine learning (ML) has transformed numerous fields, but understanding its foundational research is crucial for its continued progress. This paper presents an overview of the major classical ML algorithms and examines the state-of-the-art publications, spanning twelve decades, through an extensive bibliometric analysis study. We analyzed a dataset of highly cited papers from prominent ML conferences and journals, employing techniques such as citation and keyword analyses to uncover key insights. The study further identifies the most influential papers and authors, reveals the evolving collaborative networks within the ML community, and pinpoints prevailing research themes and emerging areas of focus. Additionally, we examine the geographic distribution of highly cited publications, highlighting the leading countries in ML research. This study provides a comprehensive overview of the evolution of traditional learning algorithms, and their impacts, and discusses challenges and opportunities for future development, with a particular focus on the Global South. The findings from this paper offer valuable insights for both ML experts and the broader research community, enhancing understanding of the field's trajectory and its significant influence on recent advances in learning algorithms.

## 1. Introduction

Machine Learning (ML), which is a subfield of artificial intelligence (AI), has exponentially grown into a transformative force, changing a multitude of industries and fundamentally altering the way we approach and solve complex real-world problems. Over the past few decades, the field of ML has evolved at an exponential pace, and its impact on our society is undeniable. ML algorithms and their variants implementation techniques have become integral to a wide range of applications, from healthcare, finance, engineering, and manufacturing to recommendation systems, autonomous vehicles, and natural language processing [1, 2]. This transformative technology has not only reshaped industries and several government critical sectors but has also changed the way we perceive and interact with our increasingly data-driven world [3].

Moreover, the rapid progress in ML has been driven by a plethora of influential publications, ranging from groundbreaking publications to innovative inventions by inspiring authors. This progress has fostered collaborative networks that have spurred diverse innovation globally. The foundational knowledge encapsulated within these classic ML publications has provided the building blocks for contemporary research and technological advancements [4, 5, 6]. Understanding the landscape of classic ML publications is thus crucial in appreciating the origins, evolution, and current state of the field, as well as charting its future trajectory.

In this context, a brief background narrative and bibliometric review study are presented in the current paper, by surveying influential early publications in ML research recognized for groundbreaking contributions made in the domain's early development. More so, our study further probes into the accumulated rich work of classic ML publications to analyze their impact, identify influential authors and papers, explore collaborative networks, dissect research themes, and investigate country-wise geographic trends. By conducting an exhaustive assessment of the existing classic ML literature, we aim to provide a holistic view of the field's historical underpinnings and its continued relevance in today's ever-evolving domain of AI and ML research and applications.

The dataset underpinning our analysis comprises the most-cited papers from reputable ML conferences and journals, spanning several decades, meticulously curated from the SCI-EXPANDED database by Clarivate Analytics [5, 7]. We employ a diverse set of bibliometric techniques, including citation analysis, co-authorship analysis, keyword analysis, and examination of publication trends, to uncover invaluable insights into the niche areas of ML research publications. Through this meticulous approach, we aim to answer pressing questions and provide a comprehensive resource for domain researchers and ML enthusiasts who wish to further investigate or analyze the rich history of ML research's exponential growth in recent years, along with its diverse groundbreaking applications.

We believe that the study findings will not only identify crucial contributions but will also reveal the web of relationships that have shaped the ML community over time. Moreover, we will illuminate the evolution of research themes and the emergence of new trends within the field, thereby promoting a deeper understanding of the ever-expanding horizons of ML research and applications. Additionally, our investigation into the geographic distribution of highly cited publications will spotlight the regional powerhouses driving innovation in the ML domain. Finally, the core ideas and technical contributions of this study can be summarized as follows:

- Perform a comparative analysis of publications, authors, and institutions historically associated with breakthroughs in classic ML.

- Quantitative assessment of the impact and prominence of classical ML studies through bibliometric indicators. This includes calculating metrics like citations, h-index, and citation velocity over time to highlight pioneering contributions.

- Developing visualizations to gain new qualitative insights into the spread of ideas and social structures within nascent ML communities.

- Classification of major theoretical foundations and algorithmic advances introduced in early ML literature. Categorizing influential papers by concept/technique and mapping their relationships and historical narrative of the field's development.

The remainder of the paper is structured as follows: Section 2 presents a taxonomical and brief overview of selected classical and foundational ML algorithms, highlighting novel designs and applications of these representative algorithms. Section 3 discusses specific insights into the challenges and limitations of the classical ML algorithms, along with their notable prospects. Section 4 provides a comprehensive analysis report of the bibliometric study conducted citing instances of highly cited relevant articles within the domain of ML research. Finally, Section 5 concludes the study.

## 2. Classical Machine Learning Algorithms Overview

Machine learning as a distinct field began to take shape in the mid-20th century with Alan Turing first introducing the concept of a machine that could learn, in his seminal paper "Computing Machinery and Intelligence" delivered in the 1950s [8]. He proposed the Turing Test as a measure of machine intelligence. This was closely followed by the work of Arthur Samuel in 1952 who developed one of the first computer programs capable of learning, a checkers-playing program that improved its performance over time through a process of learning from experience [9]. Furthermore, in 1957, Frank Rosenblatt invented the perceptron, an early type of neural network capable of learning to classify input patterns [10]. These early works laid the foundation for ML, establishing key concepts and approaches that have evolved significantly over the decades. The field has grown and diversified immensely, particularly with the advent of more advanced algorithms and the increase in computational power.

From a broader contemporary perspective, ML, a branch of artificial intelligence, utilizes various algorithms to enable computers to perform intelligent tasks. These algorithms fall into several categories: classification, regression, clustering, dimensionality reduction, ensemble methods, anomaly detection, and association rule learning. In this paper, we refer to the earliest learning algorithms as classical ML algorithms. The wide utilization of these algorithms to solve a plethora of problems in different domains has revealed their strengths and weaknesses and their suitability for domain-specific problems. Figure 1 illustrates the taxonomical representations of the aforementioned algorithm classifications. In the subsequent subsections, we briefly discuss several foundational classical ML algorithms that have led to various enhancements and the development of more advanced hybrid ML algorithms used today.

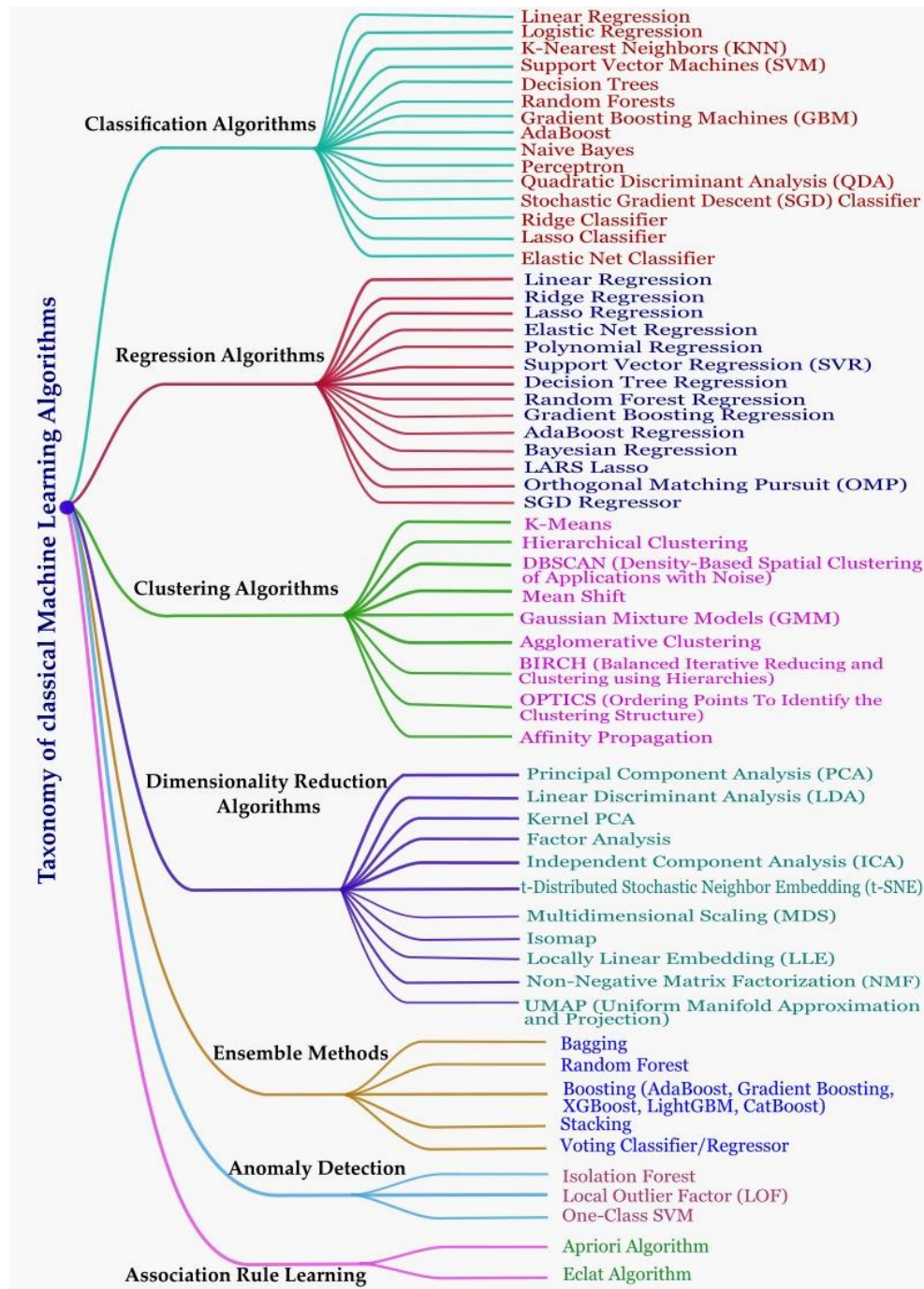

**Figure 1.** Taxonomy of ML algorithms

## 2.1. Decision Trees

Knowledge acquisition for identifying items or grouping similar items has gained considerable attention in the ML research community. Given a predefined set of classes, the goal is to select the most appropriate class for a given item, utilizing knowledge to perform classification or regression tasks as needed. A decision tree operates like a flowchart, with a rooted and directed structure. It begins at the root node progressing through decision nodes and ending at the terminal nodes. Each internal node represents a decision point, partitioning the data based on certain attributes, while each leaf node represents a class label prediction. To classify a data item, the tree is traversed from a root node according to splitting rules. These rules divide the data item's attributes within a specific domain until the item reaches a terminal node. At decision nodes, the tree can be further pruned until a terminal node is encountered. The terminal node is where decisions are finalized, as all production rules of the tree are applied by this stage. The transition from the parent node to the internal nodes and eventually to terminal or leaf nodes converts the decision tree into a set of production rules.

As a popular ML algorithm, decision trees have been found suitable for performing regression tasks or classification tasks. The suitability of decision trees for data classification has been reported by Lee et al. [11]; Sarker [12]; Klimas et al. [13]. As an ML tool, it is intuitive and hierarchical in data structure with ease of use and interpretation. The hierarchical nature of this ML model has seen its early applications in various areas with significant performance. Decision tree applications to solving problems in many real-world systems include classification tasks [14] such as in healthcare for data processing problems, diagnosis of diseases prediction [15], prompt response applications in emergencies [16], [17], [18], risk level analysis [19], traffic monitoring in a large-scale Internet of Things network [20], anomaly detection in a computer-based assessment [21], etc. The high dimensional space associated with modern real-world problems has rendered decision trees inefficient unless integrated with other algorithms.

As a foundational ML algorithm, early decision tree algorithms were characterized by several weaknesses, such as overfitting and underfitting when applied to small datasets, high sensitivity to variations in training data, poor ability to understand complex relationships, bias towards dominant classes in imbalanced distributions, and an inability to handle linear relationships among features. However, advancements in decision tree algorithms have significantly addressed these limitations, contributing to modern ML transformation. Furthermore, improvements in decision tree algorithms include the use of more sophisticated criteria for splitting, such as Gini impurity and variance reduction, instead of earlier methods like ID3. Additionally, binary search techniques have been introduced for more efficient handling of continuous variables, and methods for handling missing values during training and testing have been developed, rather than addressing them only during preprocessing as in early algorithms. The emergence of pruning techniques has also helped mitigating overfitting and improving the algorithm's generalization capabilities.

Decision trees have advanced to form the foundation of ensemble models, which involve training multiple decision trees on subsets of the original data. The inherently self-explanatory nature of decision trees is fueling the development of explainable and interpretable artificial intelligence. Modern advancements in decision tree techniques have led to improved visualization, feature importance ranking, and model interpretation. This evolution is significantly enhancing trust in modern AI solutions and improving user experiences. Personalized and recommender systems benefit from decision trees by leveraging feature importance rankings to create systems tailored to users' preferences.

## 2.2. Random Forests

Random Forest (RF) is an ML algorithm that creates a more accurate model than a single decision tree by averaging the predictions of multiple models. It combines less predictive models to achieve high performance. In RF, this merging of multiple tree models is referred to as an ensemble of trees. The development of techniques for growing multiple tree models and averaging their votes to determine the class with the highest vote has significantly improved classification accuracy in RF tasks. A RF classifier [22] is an ensemble ML method that builds a random ensemble of trees for various classification tasks. The foundational idea of RF is to use multiple data samples to construct many decision trees. The intuition behind RF is that for each input, the corresponding output tends to be similar to the outputs of similar new inputs. This is in contrast to supervised learning, which functions based on sample data.

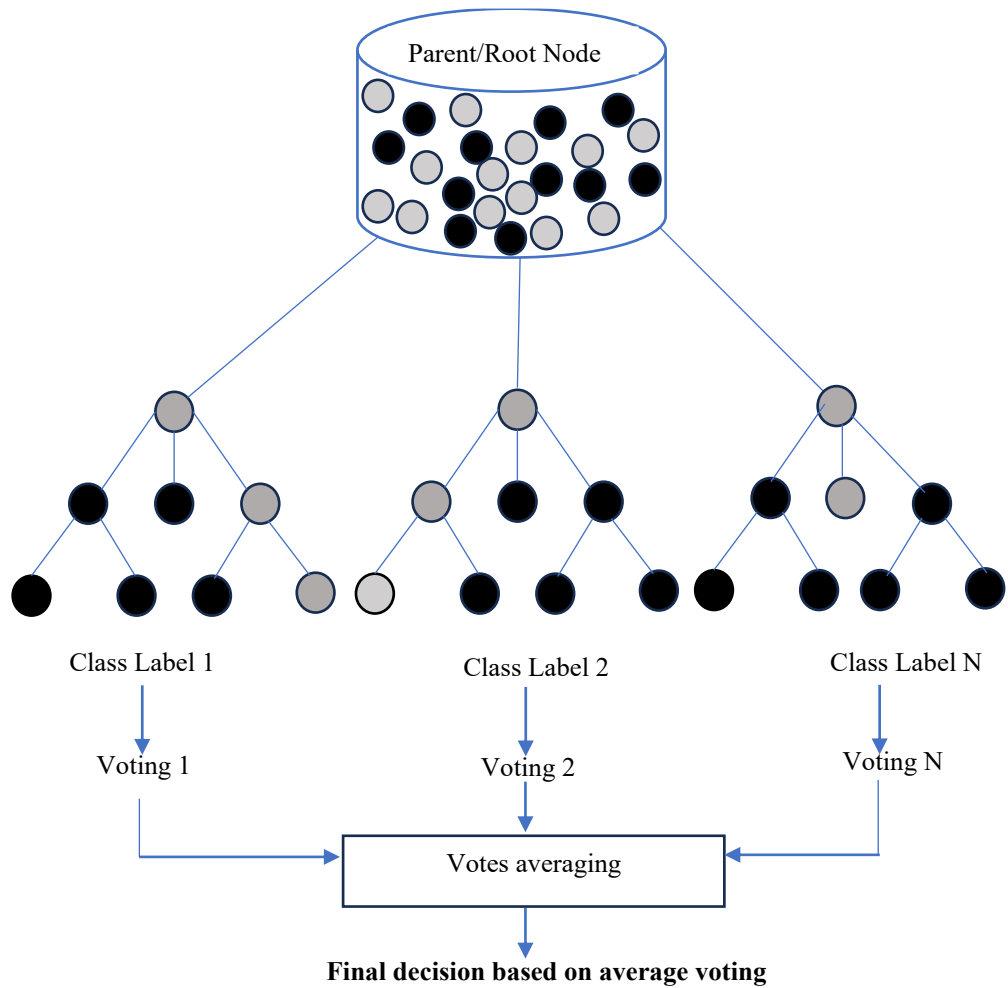

**Figure 2.** Decision Tree Classifiers for RF

In RF, several decision tree classifiers are employed in parallel assembling, as shown in Figure 2. These classifiers work on sub-samples of different data and use averaging or majority voting to decide the final result. This process significantly minimizes overfitting and reasonably increases prediction accuracy. The use of multiple decision models in RF learning algorithms results in higher accuracy compared to conventional decision tree-based models [23]. Bagging, introduced by Breiman [24], trains multiple decision trees with random feature selection [25] to build a series of decision trees, reducing variation among the original data.

Various authors have utilized RF algorithms to solve problems across different domains with significant success. RFs have been applied in educational mining [26], spatial urban data mining [27], software project management [28], clinical risk survival analysis [29, 30], predictive critical infrastructure maintenance [31], defective product prediction in manufacturing industries [32], and the analysis of large volumes of sensor data [33].

RF emerged as an ensemble method that builds multiple trees for training subsets of the original data. However, as the decision tree grows, the algorithm can become slow, and variation in features can lead to overfitting. To overcome the slow training time due to multiple trees, modern parallel computing architectures have been utilized for concurrent training. In RF, averaging the predictions of multiple decision trees provides higher accuracy than classical decision trees, making them suitable for handling noisy data and outliers. As ML algorithms advance, RF has been enhanced to support class weight balancing for datasets with imbalanced distributions.

The evolution of RF has led to incremental learning, where new trees can be introduced to the existing forest without retraining the entire model. This is significant for the success of pretraining models widely used in the AI community. In low-computing resource environments, RF models can be compressed for efficient deployment without compromising accuracy. In modern intelligent computing, RFs have been applied to various tasks. For example, in financial institutions, RF can assess credit risks and analyze financial transactions to detect suspicious activity; in the health sector, it can analyze symptoms for disease diagnosis. The evolution of RF has preserved its relevance in modern intelligent computing, maintaining its importance as a classical ML algorithm.

**2.3. Naïve Bayes**

Operationally, many ML problems appear in the form of "given A, what is B?". This involves a probabilistic model that can handle uncertainties in the sample space by determining the probabilities of outcomes [34, 35]. The Naïve Bayes classification algorithm is named after Thomas Bayes (1702-1761), who originally conceived the Bayes Theorem [36]. The goal of this method is to create a network that captures the dependencies among the variables of a given data sample. It learns the mapping of input and output parameters to predict the output of a new input. As a simple supervised ML method, it has been used to solve both binary and multi-class classification problems, such as predicting the risk of disease infection [37], spam filtering in network environments [38], decision support systems [39], and scheduling [40].

From the perspective of algorithm complexity, high-dimensional features can be easily learned by the Naïve Bayes classifier from limited training data compared to other algorithms. Naïve Bayes can perform well on small data samples for training, unlike more sophisticated techniques. However, the assumption of conditional independence does not apply in most real-world situations. Additionally, Naïve Bayes performs poorly in generalization with imbalanced datasets. Zero class probability is another notable limitation, occurring when a variable in the test data for a particular class is not observed in the training data.

The limitation of assuming conditional independence can be overcome by extending the structure to explicitly represent dependencies among attributes. In the early development of Naïve Bayes, computational resources were limited, and the algorithms focused primarily on statistics and decision theory for small-scale applications. As ML advanced, Naïve Bayes algorithms were improved to support automated decision-making in areas such as natural language processing and document classification. Modern advances have led to several variants of Naïve Bayes, extending its applications to handling binary feature vectors and integration with techniques such as semi-supervised learning. Naïve Bayes is continually being improved to handle larger datasets and implement distributed computing.

**2.4. K-Nearest Neighbors**

KNN is an unsupervised ML algorithm that is instance-based and does not generalize [41]. KNN stores instances corresponding to training data in n-dimensional space, which has been used to classify new data samples based on similarity measures such as the Euclidean distance function. KNN uses a majority vote from the KNN of each point to perform classification [42]. In this algorithm, the k parameter determines the number of neighbors chosen during the classification task. Selecting the correct k parameter positively influences the overall performance of the algorithm. KNN is efficient in handling data samples with noise.

Figure 3 shows a two-dimensional plot with two characteristics that define the class to which a given new item can belong. However, there may be more predictors with an extended number of characteristics. For example, grains are neither sweet nor crunchy, while fruits are generally sweeter than vegetables. The task is to determine the category of tomatoes (the new item). In this illustration, the four nearest food items are corn, cucumber, pawpaw, and watermelon. Since the fruits class wins the most votes, tomatoes are assigned to the fruits class.

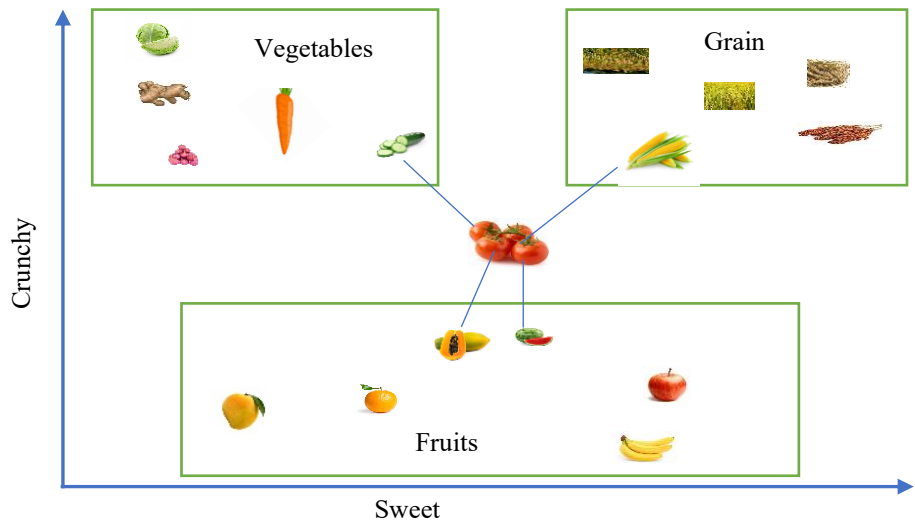

**Figure 3**. Illustration of KNN Working Principles

As shown in Figure 3, sweetness and crunchiness can be used to distinguish vegetables, grains, and fruits. In KNN, if the size of k is large, the variance impact from random error will reduce, but there is a risk of ignoring important small patterns. Choosing appropriate k-values creates a balance between overfitting and underfitting during the classification task. During classification, KNN uses majority votes from the neighbors to predict a class for a new item. In regression tasks, it uses the computed mean of the neighbors of the target variable to estimate the output value [43].

Scholars have leveraged the ease of design and adaptivity of the KNN algorithm to solve many classification and regression tasks. KNN has gained wide applications in modern society [44], including text classification [45], big data analytics [46, 47], anomaly detection in computer network log data [48], and predictive maintenance of critical industrial infrastructure [49]. However, choosing the optimal number of neighbors during classification or regression tasks is a major limitation of KNN.

The foundation of KNN is based on the principle of similarity or proximity within data samples in dimensional space. This principle has evolved with the adaptation of KNN algorithms, thanks to efficient nearest neighbor searching algorithms such as K-dimensional trees and locality-sensitive hashing, which improve computational efficiency. These new searching algorithms enhance the speed of KNN in handling large datasets. In the transition from the foundational algorithm to the present, new techniques have been developed to address imbalanced data distributions, introduce weighted voting schemes that allow closest neighbors to have more influence on predictions than farther neighbors, and achieve parallelization on multi-core processors and distributed computing architectures. With parallelization, training and prediction times on large datasets become faster.

Hybrid models have been developed by combining KNN with other algorithms to improve performance. The performance of the K-nearest neighbor has been optimized to accelerate the learning process and achieve higher accuracy. Creating clusters within each class, removing insignificant attributes, and computing similarity measures are key factors in improving KNN model performance for domain-specific tasks. In deep neural models, KNN has been applied to retrieve the nearest neighbors in large datasets, enhancing transparency in applications that require explanation. These applications contribute to the development of deep learning solutions in modern computing. Developing variants of KNN will be a promising approach to address its limitations in solving many real-world classification tasks.

## 2.5. Support Vector Machines

Support Vector Machine (SVM) is a supervised ML technique used to explore the relationships between independent and dependent variables, classify labels, and perform similar tasks [50]. In classification or regression tasks, an SVM designs a decision boundary or set of decision boundaries in n-dimensional space. The decision boundary that is farthest from the nearest training data points in each class represents a strong separation from other classes and minimizes generalization error as the margin increases. SVM effectively handles data samples in high-dimensional spaces, and the output depends on the defined mathematical functions, also referred to as kernels. However, noise in the data samples, such as overlapping target classes, can result in low performance of SVM in certain tasks.

In a given data sample, the goal of the support vector machine is to find the hyperplane that optimally differentiates clusters of vectors by categorizing the target variable on one side of the decision boundary and the other category on the opposite side of the plane [51]. The feature set that describes a class decision is called a vector, and the nearest vectors to the hyperplane are identified as the support vectors. Figure 4 shows an overview of the SVM process.

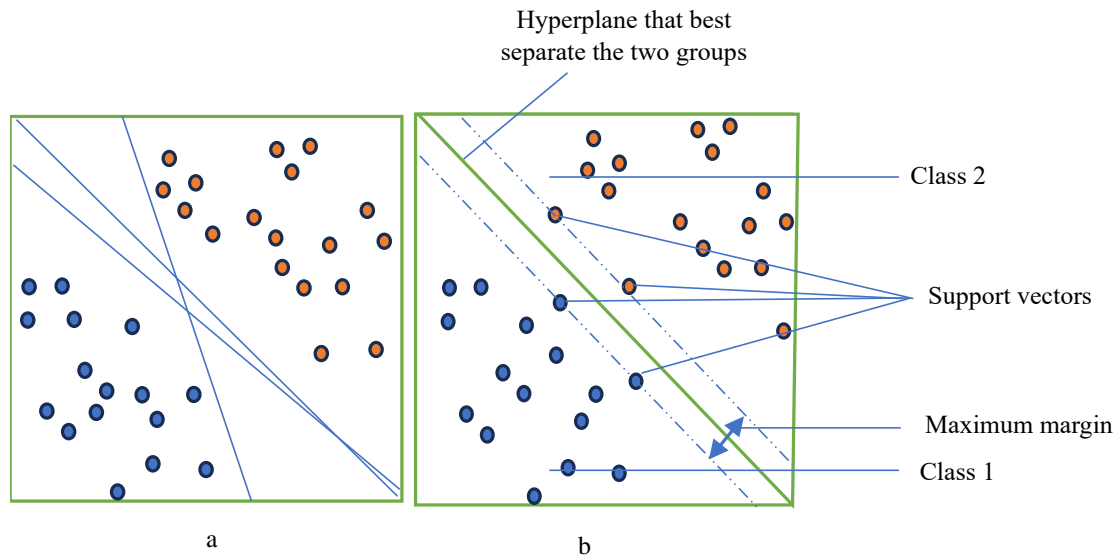

**Figure 4**. Support Vector Machines

As shown in Figure 4, (a) represents two linearly separated classes where an infinite number of hyperplanes exist, while (b) shows the distance between the decision boundaries and the closest points to the hyperplane, which defines the decision function of the SVM classifier. The hyperplanes are represented by dashed lines, and the distance between the two hyperplanes, as shown in Figure 4, is the margin. The margin determines which samples should be included in the decision function of the model. A sample can only be included if it is outside the margin and misclassified or is outrightly inside the margin. These samples are referred to as the support vectors.

There are three stages of support vector machine analysis: selection of appropriate features, training and testing, and performance evaluation of the classifier. At the feature selection stage, a new set of features is generated from the raw data to serve as input to the SVM. Most feature selection approaches follow an established procedure to show their level of significance in ranking the features [52]. Embedded, filter, and wrapper methods are commonly used for feature selection [51].

During the training and testing stage, the SVM classifier is trained using labeled observations and utilizes pre-existing information about the input and output variables to predict new label assignments. The accuracy of the SVM classifier depends on selecting the correct values of the hyperparameters. Hyperparameter values are typically set before the learning process starts and influence the value of the decision function. Generally, there are fewer hyperparameters in SVM than in other ML classification algorithms [53]. Specificity, sensitivity, and accuracy are commonly used to measure the performance of SVM, providing insights into the accuracy and reproducibility of the SVM hyperplane that differentiates classes.

Applications of SVM include text classification, pattern discovery in large data sets [54], intrusion detection [55], attack mitigation [56], classification of various diseases in the medical field [57, 58, 59, 60, 61, 62; 63, 64, 65], and other image processing applications. While large data samples and big data enhance the performance of many traditional statistical and ML algorithms, such conditions are not always met in many data analysis tasks. SVM algorithms can address these problems in various data analysis domains with limited data samples [66].

Some limitations of SVM include difficulty in obtaining the accurate value of k, especially in high-dimensional spaces, biased predictions in datasets with imbalanced distributions, and sensitivity to noisy data and outliers.

Support vector machines have evolved from theory to practice. In the early development of SVM, the concept of structural risk minimization was introduced during learning. As the algorithm advanced, the position of the hyperplane was investigated for performing binary classification tasks. Initially, the SVM algorithm was suitable only for linear classification problems. However, with the introduction of kernel functions, SVM can now handle non-linear classification problems, enabling it to address complex relationships in high-dimensional spaces. Further improvements in the SVM algorithm have allowed for multi-class classification. In deep neural models, SVM has been utilized in developing ensemble methods for feature extraction. The decision boundaries of SVM contribute significantly to building explainable and interpretable AI, which is crucial in applications where understanding the reasons behind every classification task is important, such as in medical diagnosis. The applicability of SVM can be extended by enhancing the algorithm to handle noisy data and improve performance in semi-supervised and unsupervised learning problems.

**2.6. Logistic Regression**

Logistic Regression (LR) is a statistical or probabilistic model that has been extensively explored for performing classification tasks in ML [67]. The integration of multiple classifiers, such as RF and AdaBoost, to improve classification tasks makes LR an ensemble ML technique. LR critically evaluates the relationship between different dependent and independent variables. It is primarily applied to problems where the expected output is binary, such as 1 (Yes, True, pregnant, on, available, etc.) or 0 (No, False, non-pregnant, off, not available, etc.). The basic principle of LR is to train the model by adjusting the weights of the classifiers during each boosting cycle to accurately identify the target class of a given new data item [68].

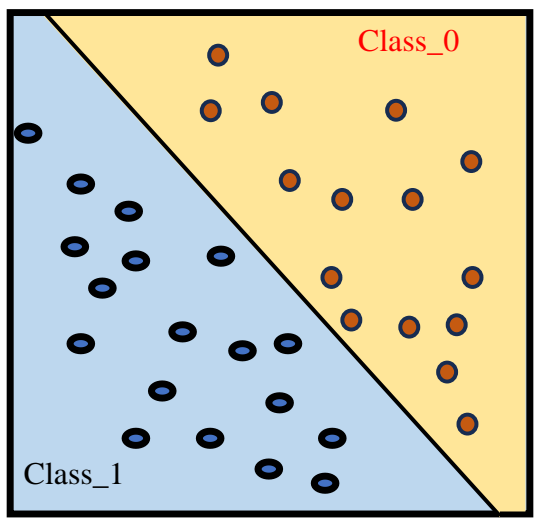

**Figure 5**. A two-dimensional linear LR model.

Logistic regression can effectively handle linearly separated data samples with high dimensionality. As shown in Figure 5, the linear decision function of LR separates the plane with a line, creating two classes. LR has been explored in various areas such as intrusion detection systems in computer networks [68], prediction of suspicious transactions in mobile money transfers [69], and trend forecasting to minimize sales uncertainty [70]. However, LR's assumption of a linear relationship between the dependent and independent variables and the decision boundaries is a notable weakness [71]. This limitation means the model cannot perform well on data samples with non-linear relationships and is sensitive to outliers. Despite this, LR has advanced from binary classification to serve as a baseline model for ensemble methods and stacking techniques. Initially, the sigmoid function was used to model the binary outcome. As the algorithm evolved, its application extended to different types of categorical data, yielding multinomial or ordinal outcomes.

The rise in computational resources and the growth of ML have influenced the development of gradient descent algorithms to optimize LR performance. During its development, regularization techniques were introduced to minimize overfitting and enhance the algorithm's ability to generalize and function in high-dimensional spaces. LR remains an integral part of the modern ML toolbox, continuing to be relevant as a baseline model for binary classification problems. It can also be combined with other complex models to build ensemble models, which can overcome data imbalance using the adaptive synthetic sampling approach.

## 2.7. Linear Regression

Linear regression can be described as a linear combination of features of different regression models. It is a simple supervised ML model that evaluates one or more inputs to predict a new value. In this regression approach, each feature possesses a coefficient that corresponds to its relative weight compared to the other features [67]. Linear regression can be either simple or multiple, depending on the data samples [72]. The application of linear regression is relevant in any domain where a linear relationship between two quantitative variables (correlation between data features and the target variables) can be easily established [73, 74]. This is evident in the wide application of linear regression for prediction tasks in many health-related interventions [75], exploration of financial and non-financial variables to forecast fraud in the financial sector [76], analysis of behavioral patterns, modeling of disease progression, and more.

In complex and large-scale datasets, validating the linearity of variables and the independence of errors is challenging. Linear regression has evolved as an algorithm to check if there is a relationship between two given variables. The algorithm has developed from the initial stage of least squares methods to becoming part of modern ML toolboxes such as neural networks. Initially, least squares methods were used to find the minimal regression line between observed and predicted values. The subsequent formulation of matrix notation in linear regression improved the computational efficiency of the algorithm and enabled it to perform regression tasks on multiple variables.

Further advancements in linear regression include the development of statistical theory allowing hypothesis testing, confidence intervals, significance levels, generalized linear models to handle distributions of non-normal errors, and robust regression to minimize the impacts of outliers. These advancements have significantly enhanced the computational efficiency of linear regression. Modern computers and the development of statistical software applications allow the analysis of large-scale data samples using regression analysis. As algorithms advance, linear regression is combined with more complex models to build advanced modeling techniques for big data analytics and other complex computing tasks.

## 2.8. Perceptron Model

The perceptron model is one of the earliest classical ML techniques, introduced in the 1950s by Rosenblatt [77]. The model is considered the foundation basis of our understanding of artificial neural networks, which have recently transformed the ML field through the advent of deep learning and its vast real-world applications. Although the original perceptron model, with its few layers, is limited to simple logic tasks and cannot handle complex logic functions or classify linearly inseparable patterns, several studies and research efforts have enhanced it into a more robust learning variant. This transformation was notably accelerated by the introduction of the backpropagation approach used in training multilayered perceptron models [78, 79]. More so, the concept of supervised learning was first introduced with the training of the perceptron model. Supervised learning here refers to comparing the output resulting from the model training with the desired or actual output of interest based on specific objective functions. During training, the objective function aims to reach zero relative to a probability feedback loop. The perceptron model is applicable in solving both regression and classification problems. Figure 6 illustrates the simple perceptron model.

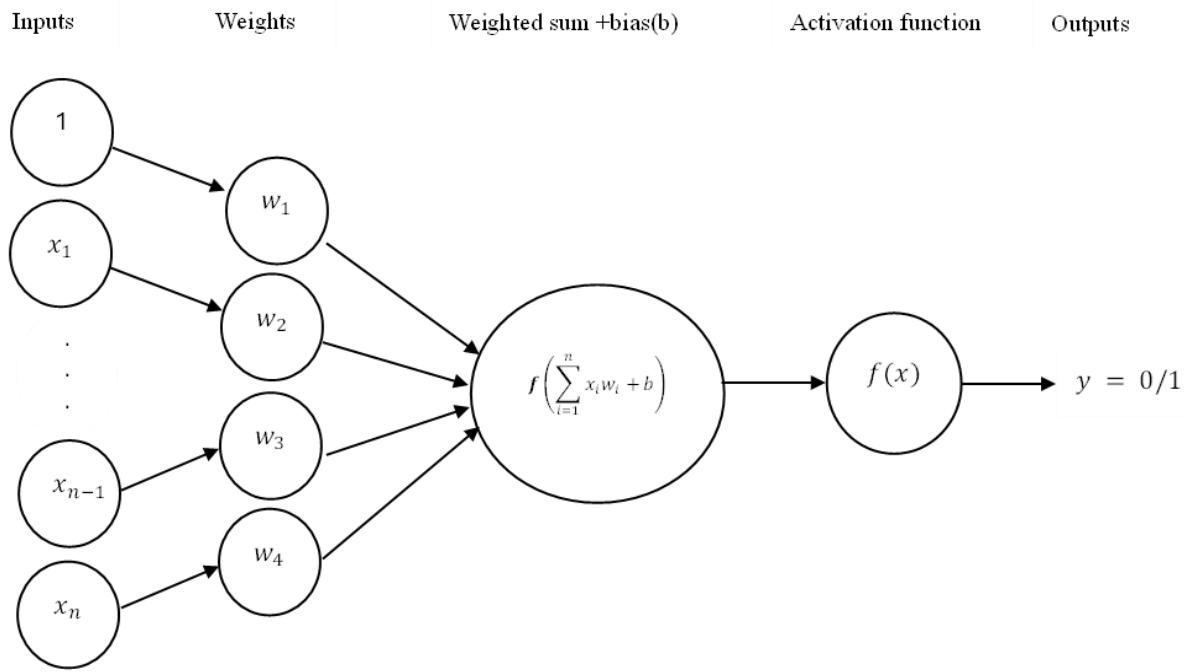

**Figure 6**. The architecture of the classical perceptron model

Despite its classic architectural model, the perceptron model has significant strengths. It is conceptually simple and easy to implement, making it a good starting point for understanding neural networks. The model is computationally efficient for linearly separable data, capable of quickly converging to a solution, and is particularly effective for handling binary classification problems. Also as earlier mentioned, the perceptron model serves as the building block for more complex neural networks such as the multilayer perceptron and deep learning networks. However, the perceptron model lacks hidden layers which limit its capabilities to capture more complex patterns and relationships in data. Furthermore, because it only produces binary outputs, it is unsuitable for problems requiring probabilistic interpretations or multi-class classification. The model's simplicity also limits its expressiveness compared to more advanced models with multiple layers and non-linear activation functions. However, despite these limitations, the perceptron model remains an important milestone in the history of ML as it continues to provide useful guidance into the principles of artificial neural networks in general. Figure 7 provides an insight into the historical evolution of early artificial intelligence research trajectories.

To address the limitations of the simple perceptron model in handling linearly separable patterns, the multilayer perceptron was introduced by Rumelhart et al. [79] and trained using the backpropagation algorithm. Baum [80] defined the multilayer perceptron as a model consisting of an input layer with $x$ units, one or more intermediate (hidden) layers, and an output layer with $y$ units. In this model, each unit in the output layer is connected to each input unit in the subsequent layer. Each connection from the $j^{th}$ unit in layer $l$ to the $i^{th}$ unit in layer $l+1$ is assigned an associated weight. For an in-depth review and explanation of the perceptron model, refer to the research conducted by Du et al. [81].

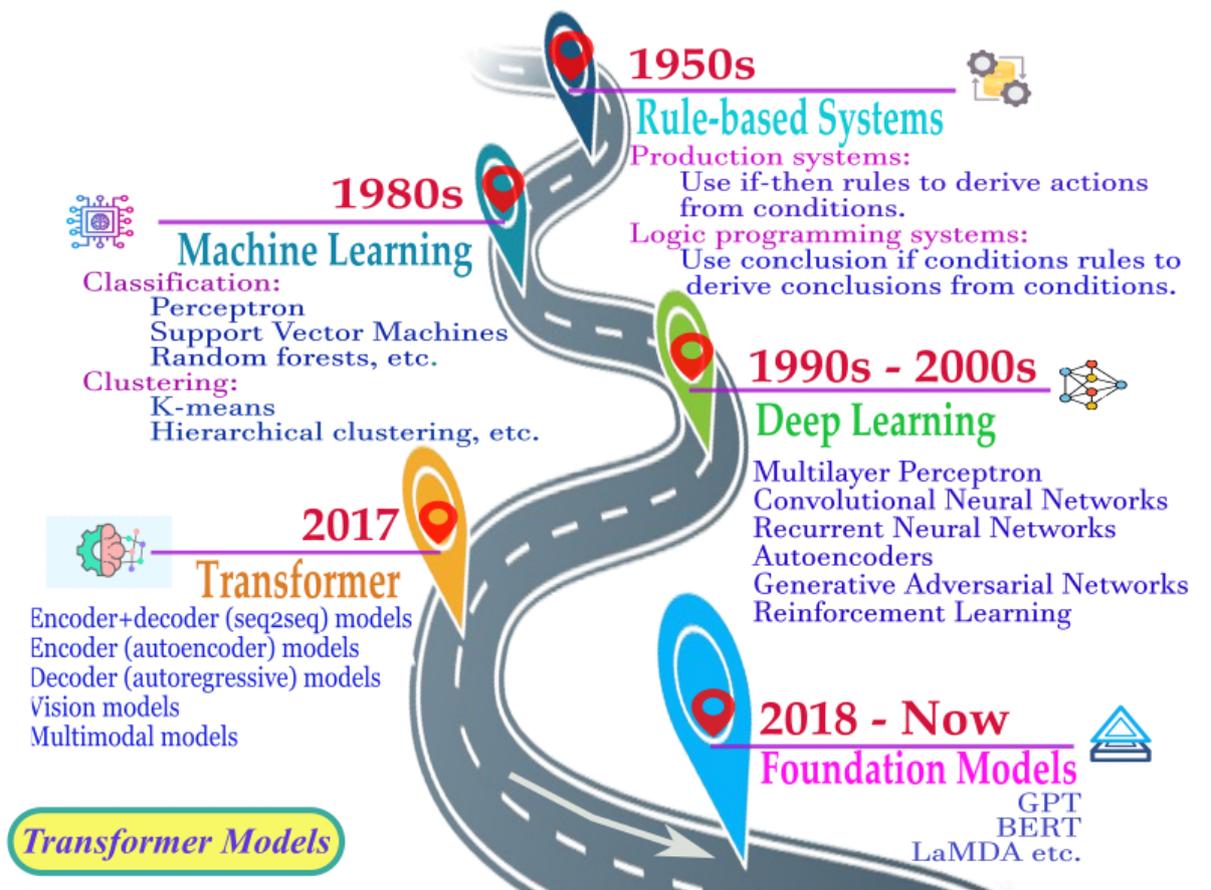

**Figure 7.** Historical timeline from early AI research to the foundation model era

## 3. Classical Machine Learning Limitations and Prospects

Classical ML algorithms have performed exceptionally well in the past. Their significant performance was anchored on the availability and accessibility of limited data with low-dimensional space and the assumption of linearity in the data model. However, the explosion of data generation driven by advancements in Internet technology has rendered these algorithms significantly inefficient without considering hybridization and the combination of algorithms to build ensemble models. Moreover, in real-world scenarios, data representations are not always linear.

The advent of deep neural networks, the high volume of data, and high-performance computing resources have inspired artificial intelligence solutions that have shifted attention from foundational ML algorithms to more sophisticated algorithms. However, these classical algorithms can still be creatively utilized as base learners to build Explainable Artificial Intelligence (XAI) systems, providing interpretation and understanding of black-box models. This approach is promising for gaining user trust in system reliability, especially in regions where AI solutions have not yet gained massive ground.

It is equally important to emphasize that ML in the Global South faces critical limitations compared to its Global North counterpart due to infrastructural, socio-economic, and technical challenges. Limited internet access and inadequate computing resources hamper data collection and processing capabilities, while economic constraints and educational barriers restrict investment in technology and local expertise development. Additionally, the scarcity of high-quality, localized datasets and the cultural and linguistic biases in existing models reduce the effectiveness of ML applications. These issues are compounded by underdeveloped regulatory frameworks and societal concerns about data privacy and ethics, collectively impeding the widespread and effective adoption of ML technologies in the region.

De-Arteaga, Herlands, Neill, and Dubrawski [82] further emphasize the importance of considering local context when designing and implementing ML projects for developing countries. The authors propose a three-step roadmap for advancing global development with ML: first, improving data reliability by addressing the challenges of incomplete, noisy, and biased data common in developing regions; second, providing direct solutions and deployed systems by developing ML-powered systems that directly address critical development issues within these regions; and third, informing policy and decision-makers by ensuring that ML insights are translated into actionable policy recommendations. The authors argue that the challenges faced in adopting and deploying ML services and their associated applications in these regions, such as limited data availability and infrastructure constraints, can be viewed as opportunities for advancing ML itself. They suggest that these challenges can motivate the development of novel ML methodologies and algorithms that are more robust, efficient, and adaptable to resource-constrained environments.

A more detailed discussion on the critical challenges faced in the development and implementation of ML algorithms in developing countries (the Global South), along with their respective approaches to addressing these challenges, will be presented in the later stages of this paper.

### 3.1. Prospects

Traditional ML models were initially designed to solve linear classification or regression tasks and later extended into non-linear environments. Their performance depends on the nature and quantity of data used. When good quality and sufficient quantities of data are available, traditional ML models perform excellently. However, they degrade in performance when data samples are available in hundreds of thousands or millions, especially when used in isolation.

To revive the involvement of traditional ML models in the present computation-challenging world, techniques like ensemble and hybridization should be thoroughly explored. ML communities should examine how best to combine two or more models by employing methods such as boosting, bagging, stacking, or hybridizing traditional ML algorithms with state-of-the-art ML models [83]. Ensemble and hybridization techniques harness the strengths of individual models to achieve better performance. For instance, similar to evolutionary and metaheuristic optimization techniques, ensembles mimic human behavior (voting), extending their capability beyond classification and regression tasks. Ensemble and hybridization are crucial to ensure the relevance of classical ML models in the field of Artificial Intelligence.

Zhang and Ma [84] reported that ensemble techniques such as boosting, consensus aggregation, classifier combination, classifier selection, bagging, and stacked generalization can be extended for deep model optimization while harnessing classical ML advantages such as effective computation speed, efficient resource utilization, and scalability. Using an ensemble approach, the layer-by-layer deep learning technique could be adopted in classical ML implementation. For example, Zhao and Feng [85] employed an ensemble learning technique and proposed a non-differentiable deep model (Deep Forest), which is a typical ensemble of decision trees. The proposed method implements an ensemble of trees at each layer and was reported to have provided good performance. Generally, ensemble techniques have shown promising results when applied to high-dimensional data tasks. Exploring ensemble techniques is a resolution for maintaining ML prominence in the field of computational intelligence.

Aside from ensembles, hybridization is another renowned ML technique capable of improving classical ML applications in the field of intelligent computation. There are different approaches to exploring the hybridization of ML models; the two prominent ones are the hybridization of two or more classical ML models and the hybridization of classical ML models with deep networks. While the optimal hybridization among classical ML models is still under exploration, the hybridization of ML algorithms and deep networks has shown promising results and should be further investigated. Recently, some classical ML algorithms like support vector machines, RFs, and decision trees have been hybridized successfully with deep learning models, generating impressive accuracy and resource utilization performance.

The hybridization of classical ML models with deep networks is a factor to consider in resolving the explainability and interpretability challenges in deep learning models. Deep learning models are regarded as black-box models because the details of their implementation are often unknown to users. The white-box models (classical ML models) with significant explainability and interpretability features could be harnessed to address these challenges. Leveraging the strengths of both deep networks and classical ML models through hybridization techniques could result in efficient and enhanced deep models with improved user interface mechanisms. Hybridization that critically leverages the benefits of both deep learning models and traditional ML models could lead to the expected revolution, preserving the relevance of classical ML in the field.

In the next section, we present a detailed exploratory analysis and discussion using bibliometric analysis techniques to gain insights into the state-of-the-art research in the evolution trajectory of machine learning (ML). We assume that this statistical analysis will enable us to track the output and impact of notable researchers within the ML domain. Additionally, we employ visualization features of the statistical analysis method to track and understand publication relations, effectively measuring the influence of ML publications in the scientific community.

## 4. Bibliometric Analysis

The SCI-EXPANDED database from Clarivate Analytics was employed to obtain the data utilized in this study (data extracted on 23 October 2023). The 2022 journal Impact Factor ($IF_{2022}$) was reported in the Journal Citation Report (JCR) on 28 June 2023. According to the definition of journal impact factor, it is best to search for documents published in 2022 in SCI-EXPANDED after the $IF_{2022}$ are released. Please note that in order to obtain an accurate bibliometric analysis report using SCI-EXPANDED, it is necessary to extract the previous year's report based on the Journal Citation Report (JCR) from the middle or end of the current year onwards. Consequently, the citation report for 2023 can only be obtained around the middle or end of 2024.

To ensure thorough search coverage, we employed the use of quotation marks (" ") and the Boolean operator "or" to guarantee the inclusion of at least one search keyword in the terms of TS (topic) including title, abstract, author keywords, and *Keywords Plus*, from 1900 to 2022. The search keywords were "machine learning". To maintain analysis accuracy, we also included fewer common terms like "machine learnings" in the SCI-EXPANDED databases. A total of 192 documents with 1,000 total citations or more were searched from 1989 to 2022. The index of citations from the Web of Science was updated as time went on. In 2011, Ho's research group proposed a citation indicator, $TC_{year}$, the total number of citations from the Web of Science Core Collection since publication year until the end of the most recent year (it is 2022 in this study, $TC_{2022}$) [86]. The advantage of $TC_{2022}$ is that it is an invariant parameter, thus ensuring repeatability, while the index of citation would have been updated from time to time [87]. $TC_{2022}$ of 1,000 or more was used to retrieve the classic documents. A total of 160 classic documents (83% of 192 documents) were found. Documents searched out just by *Keywords Plus* tend to be irrelevant to the search topic. Ho's group was the first to provide suggestions to apply the "front page" filter, which included articles' titles, author keywords, and abstracts [86]. This filter may prevent the introduction of irrelevant papers for further analysis in the bibliometric study. Finally, all the 160 classic ML research documents with search keywords on their "front page" were defined. Figure 8 illustrates the schematic for searching the classic publications in ML research.

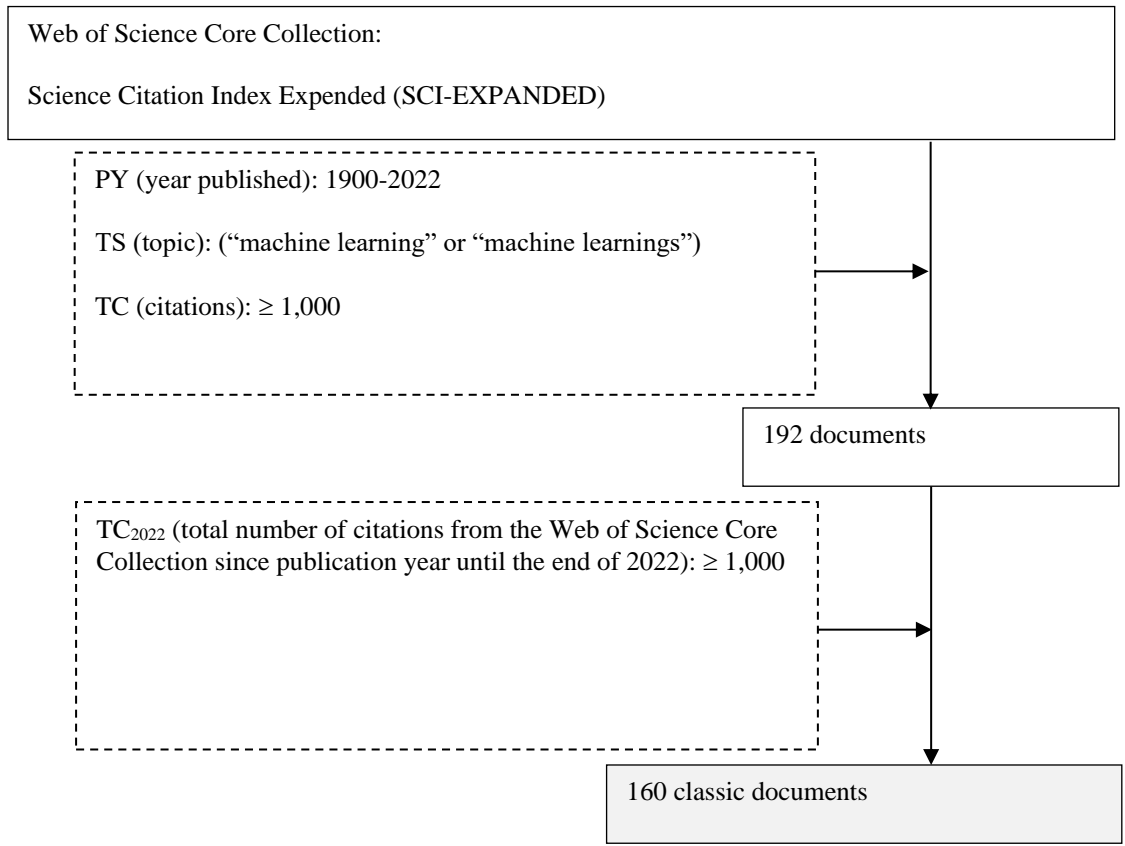

**Figure 8**. Schematic for searching the classic publications in ML research

Figure 8 distinguishes between dotted line segments, which indicate our actions, and solid line segments, which represent the outcomes of those actions. In the initial step, we searched with the following criteria: PY (year published) between 1900 and 2022, TS (topic) including either "machine learning" or "machine learnings," and TC (citations) equal to or exceeding 1,000. These operations utilized the functions from SCI-EXPANDED. As a result, we identified 192 documents. In the subsequent step, we applied a filter based on TC2022 (total number of citations from the Web of Science Core Collection since the publication year up to the end of 2022) exceeding 1,000. For this task, we employed functions from EXCEL, as detailed in the methods section. This refinement yielded 160 documents.

The whole SCI-EXPANDED record and the number of yearly citations for each document were taken and examined in Microsoft Excel 365. Then, additional code or formula was conducted by hand manually [88]. Affiliations of authors in England and Scotland were regrouped as one group under the heading of the United Kingdom (UK). In the Web of Science database, the corresponding author was designated as the "reprint author"; this study instead used the term "corresponding author". In a single-author article where authorship was unspecified, the single author was both the first author and corresponding author. In a single institution article, the institution was classified as both the first author's institution and the corresponding author's institution [89]. Similarly, for a single-country article, the country was classified as both the first author's country and the corresponding author's country. In multiple corresponding author articles, all corresponding authors, institutions, and countries are considered [90, 91].

To investigate the citations received by the ML-related publications, we used three indicators: $TC_{year}$, $C_{year}$ (the number of citations in a particular year. $C_{2022}$ means the number of citations in 2022, and $CPP_{year}$ (average number of citations per publication ($CPP_{2022} = TC_{2022}/TP$) [92].

**4.1. Results and Discussion**

The following subsections present several intriguing findings from the bibliometric analysis study, which was conducted using extensive search methods spanning 12 decades of state-of-the-art classical ML research. The analysis reveals key results from the inception of ML to the present, highlighting the most significant breakthroughs in ML algorithms and their applications. Additionally, these findings identify pioneering authors and researchers who have significantly contributed to this field, providing valuable insights for ML enthusiasts.

**4.1.1 Document type and language of publication**

A total of 160 classic documents in ML research with $TC_{2022}$ of 1,000 or more within five document types indexed in the Web of Science (Table 1). Among the document types, articles received 124 publications (78% of 160 publications) with total citations of $TC_{2022}$ = 354,582 citations and reviews had 32 (20%) publications with $TC_{2022}$ = 100,636 citations. The $CPP_{2022}$ of the document type of reviews was 1.1 times of articles. Six of 160 classic documents had $TC_{2022}$ of 10,000 or more including five articles and one review entitled "Gradient-based learning applied to document recognition" [93] by Lecun et al. from the AT&T Bell Laboratories in the USA and the University of Montreal in Canada.

**Table 1**. Characteristics of document types

| Document Type | TP | % | AU | APP | $TC_{2022}$ | $CPP_{2022}$ |
|---|---|---|---|---|---|---|
| Article | 124 | 78 | 958 | 7.7 | 354582 | 2860 |
| Review | 32 | 20 | 184 | 5.8 | 100636 | 3145 |
| Proceedings paper | 6 | 3.8 | 15 | 2.5 | 12455 | 2076 |
| Editorial material | 4 | 2.5 | 12 | 3.0 | 5036 | 1259 |
| Book chapter | 1 | 0.63 | 3 | 3.0 | 1875 | 1875 |

*TP*: number of classic publications; *AU*: number of authors; *APP*: average number of authors per publication; $TC_{2022}$: the total number of citations from Web of Science Core Collection since publication year until the end of 2022; $CPP_{2022}$: average number of citations per publication ($TC_{2022}$/*TP*).

One of the earliest significant review articles on ML is the study presented by Mitchell et al. [94], which was published in the Annual Review of Computer Science Journal. The authors provide a review of the field of ML from 1990. It outlines several grand challenges for ML research over the next decade, such as developing a learning household robot or automated discovery of important regularities in scientific databases. It summarizes recent progress in areas like inductive generalization and neural networks. It also identifies missing science at the time, including the need for improved generalization from examples and methods for incremental and modular learning. This paper from 1990 helps provide a snapshot of the state of the field early in its growth. It articulated ambitious goals that drive much of current ML research today, over 30 years later, as areas like robot learning, discovery from large datasets, and lifelong learning remain active topics that have seen partial but important progress. The challenges outlined in this paper still resonate as guides for the development of strong ML.

The earliest document type of editorial material was entitled "Neural Networks and the Bias/Variance Dilemma" [95]. This paper discusses neural networks from a statistical perspective of nonparametric regression. It frames neural network learning problems as estimating regression functions from data in a model-free manner. A key contribution is decomposing error into bias and variance, showing how nonparametric approaches have high variance even as they are consistent, requiring very large datasets. This "bias-variance dilemma" creates fundamental challenges for applications like the perception that neural networks were applied to. The paper significantly argued that incorporating careful inductive biases is necessary for success on complex tasks, shifting focus from solely learning to representation and model selection. This work was highly influential, establishing statistical foundations for understanding generalization and providing lessons still important today regarding the role of priors in deep learning for vision and language.

Moreover, the earliest document type of article entitled "A Bayesian Method for the Induction of Probabilistic Networks from Data" was presented by Cooper and Herskovits [96]. This paper presents a Bayesian method for constructing probabilistic networks from databases. The key contribution is a formula for computing the posterior probability of a network structure given a database, based on several assumptions. This allows ranking network structures by probability and finding the most probable one. The paper also discusses handling missing data, and hidden variables, and performing probabilistic inference using the learned networks. This work on inductive learning of Bayesian networks from data was pioneering and highly influential. It helped establish Bayesian network models and laid foundations for modern areas like structure learning from data, dealing with missing values, and combining multiple learned models for inference. These continue to be important research challenges in statistical relational artificial intelligence and ML.

However, it is noteworthy that only articles were used for subsequent analysis because they included complete research ideas and results. As a result, we identified 124 classic articles in the ML research, all of which were published in English.

### 4.1.2. Characteristics of publication outputs

Ho [97] proposed a relationship between the total number of articles in a year ($TP$) and their citations per publication ($CPP_{year}$) by the years in a topic. In the last decade, it has been applied as a unique indicator in SCI-EXPANDED for classic articles on a research topic, for example, *Helicobacter pylori* [98] and apoptotic [99]. The distribution of classic ML articles is displayed in Figure 9. It was generally accepted that time is needed to accumulate a total number of citations for an article. It took $TP$s about five years to reach a peak which was much longer than the case of classic *Helicobacter pylori* articles with 16 years. The results show that ML research has been a popular topic in recent years. The highest $CPP_{2022}$ was 11,039 in 2014 which can be attributed to the article by Srivastava et al. [100] with a $TC_{2022}$ of 19,859 (rank 3$^{rd}$) and $C_{2022}$ of 3,788 (rank 4$^{th}$).

### 4.1.3. Web of Science categories and journals

To gain insight into the prominent areas of ML research, we examined the distribution of Web of Science categories among classic articles. In 2022, the Journal Citation Reports (JCR) encompassed 9,537 journals with citation references spanning 178 Web of Science categories within SCI-EXPANDED. Among these, 124 classic ML articles were found in journals affiliated with all 52 of the Web of Science categories in SCI-EXPANDED. Seven of these categories stood out with 10 or more classic articles, including artificial intelligence and computer science (145 journals) with 46 classic articles, accounting for 37% of the total 124 articles; electrical and electronic engineering (275 journals) with 24 articles (19%); multidisciplinary Sciences (73 journals) with 12 articles (9.7%); theory and methods in computer science (111 journals) with 12 articles (9.7%); automation and control systems (65 journals) with 11 articles (8.9%); and biochemistry and molecular biology (285 journals) with 10 articles (8.1%).

In total 124 classic ML-related articles were published in 82 journals in SCI-EXPANDED. Eight journals published three classic articles or more such as the *Journal of machine learning Research* ($IF_{2022}$ = 6.0) with 10 articles (8.1% of 124 articles), *Nature* ($IF_{2022}$ = 64.8) (7 articles, 5.6%), *IEEE Transactions on Pattern Analysis and Machine Intelligence* ($IF_{2022}$ = 23.6) (5, 4.0%), *Machine Learning* ($IF_{2022}$ = 7.5) (4, 3.2%), and three articles (2.4%) in each of *Nature Methods* ($IF_{2022}$ = 48), *Nucleic Acids Research* ($IF_{2022}$ = 14.9), *Cell* ($IF_{2022}$ = 64.5), and *Proceedings of the IEEE* ($IF_{2022}$ = 20.6) respectively.

### 4.1.4. Countries and institutions

There were 124 classic ML articles in SCI-EXPANDED from 34 countries. A total of 74 articles (60% of 124 articles) were single-country articles from 14 countries with $CPP_{2022}$ of 3,117 citations and 50 (40%) articles were internationally collaborative articles from 31 countries with $CPP_{2022}$ of 2,478 citations. The six publication indicators and each of the related citation indicators ($CPP_{2022}$) were recently proposed to compare the country's publication performance [101]. Table 2 listed the top 11 productive countries with five classic articles or more as well as the six publication indicators and their $CPP_{2022}$. The USA dominated in all six publication indicators with a $TP$ of 79 articles (64% of 124 articles), an $IP_C$ of 44 articles (59% of 74 single-country articles), a $CP_C$ of 35 articles (70% of 50 internationally collaborative articles), an $FP$ of 60 articles (48% of 124 first-author articles), an $RP$ of 62 articles (50% of 124 corresponding-author articles), an $SP$ of nine articles (64% of 14 single-author articles). Comparing the top 11 countries in Table 2, France with a $TP$ of 10 articles, an $FP$ of three articles, and an $RP$ of three articles had the highest $CPP_{2022}$ of 4,712, 11,150, and 11,150 citations respectively. It can be attributed to the article entitled "Scikit-learn: machine learning in Python" [102] by Pedregosa as both the first author and the corresponding author from CEA Saclay in France with a $TC_{2022}$ of 30,449 (rank 1$^{st}$) and a $C_{2022}$ of 7,859 (rank 1$^{st}$). Canada with a $CP_I$ of nine articles had the highest $CPP_{2022}$ of 5,575 citations. The UK with a $CP_C$ of 10 articles had the highest $CPP_{2022}$ of 5,483 citations. The USA with an $SP$ of nine articles had the highest $CPP_{2022}$ of 4,488 citations.

**Table 2.** Top 11 most productive countries with five articles or more.

| Country | TP | TPR (%) | $CPP_{2022}$ | $IP_CR$ (%) | $CPP_{2022}$ | $CP_CR$ (%) | $CPP_{2022}$ | FPR (%) | $CPP_{2022}$ | RPR (%) | $CPP_{2022}$ | SPR (%) | $CPP_{2022}$ |
|---|---|---|---|---|---|---|---|---|---|---|---|---|---|
| USA | 79 | 1 (64) | 2,625 | 1 (59) | 2,656 | 1 (70) | 2,585 | 1 (48) | 2,390 | 1 (50) | 2,358 | 1 (64) | 4,488 |
| Germany | 16 | 2 (13) | 3,591 | 5 (4.1) | 1,593 | 2 (26) | 4,052 | 2 (10) | 1,870 | 2 (10) | 1,870 | 3 (7.1) | 2,009 |
| Canada | 15 | 3 (12) | 3,074 | 2 (8.1) | 5,575 | 5 (18) | 1,407 | 3 (7.3) | 4,266 | 3 (7.3) | 4,266 | 2 (14) | 4,383 |
| UK | 15 | 3 (12) | 4,103 | 3 (6.8) | 1,341 | 3 (20) | 5,483 | 4 (5.6) | 2,353 | 5 (5.6) | 2,353 | N/A | N/A |
| France | 10 | 5 (8.1) | 4,712 | N/A | N/A | 3 (20) | 4,712 | 8 (2.4) | 11,150 | 8 (2.4) | 11,150 | N/A | N/A |
| China | 10 | 5 (8.1) | 2,505 | 3 (6.8) | 3,578 | 8 (10) | 1,432 | 4 (5.6) | 2,866 | 4 (6.5) | 2,656 | N/A | N/A |
| Australia | 9 | 7 (7.3) | 2,656 | 8 (1.4) | 3,693 | 6 (16) | 2,526 | 6 (3.2) | 3,357 | 6 (4) | 3,015 | 3 (7.1) | 3,693 |
| Netherlands | 8 | 8 (6.5) | 2,097 | N/A | N/A | 6 (16) | 2,097 | 8 (2.4) | 1,396 | 8 (2.4) | 1,396 | N/A | N/A |
| Spain | 7 | 9 (5.6) | 1,489 | 6 (2.7) | 1,466 | 8 (10) | 1,498 | 6 (3.2) | 1,473 | 7 (3.2) | 1,473 | N/A | N/A |
| Switzerland | 5 | 10 (4) | 2,199 | 8 (1.4) | 2,419 | 11 (8) | 2,144 | 11 (1.6) | 1,751 | 11 (1.6) | 1,751 | N/A | N/A |
| Italy | 5 | 10 (4) | 1,603 | N/A | N/A | 8 (10) | 1,603 | 12 (0.81) | 3,160 | 12 (0.81) | 3,160 | N/A | N/A |

*TP*: total number of articles; *TPR* (%): rank of total number of articles and percentage; $IP_CR$ (%): rank of single-country articles and percentage in all single-country articles; $CP_CR$ (%): rank of internationally collaborative articles and percentage in all internationally collaborative articles; *FPR* (%): rank of first-author articles and percentage in all first-author articles; *RPR* (%): rank of corresponding-author articles and percentage in all corresponding-author articles; *SPR* (%): rank of single-author articles and percentage in all single-author articles; $CPP_{2022}$: average number of citations per publication ($TC_{2022}/TP$); N/A: not available.

Table 3 lists the top 10 productive institutions with four classic articles or more with the five publication indicators and their $CPP_{2022}$ [101]. Among these 10 institutions, seven of them derived from the USA and one from each of Australia, Canada, and Spain respectively. Stanford University in the USA ranked the top in the four publication indicators with a *TP* of eight articles (6.5% of 124 articles), an $IP_I$ of two articles (5.1% of 39 single-institution articles), an *FP* of five articles (4.0% of 124 first-author articles), and an *RP* of five articles (4.1% of 123 corresponding-author articles). The Massachusetts Institute of Technology (MIT) ranked the top in the two publication indicators with a *TP* of eight articles (6.5% of 124 articles) and a $CP_I$ of eight articles (9.4% of 85 inter-institutionally collaborative articles). Comparing the top 10 institutions in Table 3, the University of Toronto in Canada with a *TP* of four, an $IP_C$ of one, an *FP* of two, and an *RP* of two articles, had the highest $CPP_{2022}$ of 6,102, 19,859, 11,025, and 11,025 citations respectively. The University of Washington in the USA with a $CP_C$ of seven articles, had the highest $CPP_{2022}$ of 5,687 citations.

**Table 3.** Top 10 productive institutions

| Institution | TP | TPR (%) | CPP$_{2022}$ | IP$_I$ IP$_I$R (%) | CPP$_{2022}$ | CP$_I$ CP$_I$R (%) | CPP$_{2022}$ | FP FPR (%) | CPP$_{2022}$ | RP RPR (%) | CPP$_{2022}$ |
|---|---|---|---|---|---|---|---|---|---|---|---|
| MIT, USA | 8 | 1 (6.5) | 1,379 | N/A | N/A | 1 (9.4) | 1,379 | 2 (3.2) | 1,390 | 2 (3.3) | 1,390 |
| Stanford Univ, USA | 8 | 1 (6.5) | 2,448 | 1 (5.1) | 2,432 | 3 (7.1) | 2,453 | 1 (4.0) | 2,168 | 1 (4.1) | 2,168 |
| Univ Washington, USA | 7 | 3 (5.6) | 5,687 | N/A | N/A | 2 (8.2) | 5,687 | 5 (1.6) | 1,280 | 2 (3.3) | 1,283 |
| Harvard Univ, USA | 6 | 4 (4.8) | 2,541 | N/A | N/A | 3 (7.1) | 2,541 | 17 (0.81) | 1,766 | 13 (0.81) | 1,766 |
| Univ Calif San Francisco, USA | 6 | 4 (4.8) | 1,841 | N/A | N/A | 3 (7.1) | 1,841 | 17 (0.81) | 1,116 | 13 (0.81) | 1,116 |
| Univ Calif Berkeley, USA | 5 | 6 (4.0) | 4,448 | 4 (2.6) | 15,074 | 7 (4.7) | 1,791 | 3 (2.4) | 6,023 | 4 (2.4) | 6,023 |
| Univ Melbourne, Australia | 5 | 6 (4.0) | 2,984 | N/A | N/A | 6 (5.9) | 2,984 | 5 (1.6) | 3,689 | 6 (1.6) | 3,689 |
| Univ Toronto, Canada | 4 | 8 (3.2) | 6,102 | 4 (2.6) | 19,859 | 10 (3.5) | 1,516 | 5 (1.6) | 11,025 | 6 (1.6) | 11,025 |
| Massachusetts Gen Hosp, USA | 4 | 8 (3.2) | 2,764 | N/A | N/A | 7 (4.7) | 2,764 | N/A | N/A | N/A | N/A |
| Univ Granada, Spain | 4 | 8 (3.2) | 1,473 | N/A | N/A | 7 (4.7) | 1,473 | 5 (1.6) | 1,491 | 6 (1.6) | 1,491 |

*TP*: total number of articles; *TPR* (%): the rank and the percentage of total articles in the total number of articles; *IP$_I$R* (%): the rank and the percentage of single-institution articles in the total single-institution articles; *CP$_I$R* (%): the rank and the percentage of inter-institutionally collaborative articles in the total inter-institutionally collaborative articles; *FPR* (%): the rank and the percentage of first-author articles in the total first-author articles; *RPR* (%): the rank and the percentage of the corresponding-author articles in the total corresponding-author articles; *CPP*: number of citations per publication ($TC_{2022}/TP$); N/A: not available.

### 4.1.5. Citation histories of the most frequently cited articles

The total number of citations ($TC_{2022}$) is an indication of an article with high impact or visibility in the research community. However, an article's impact might not always be high immediately after its publication. Therefore, the number of citations in the most recent year ($C_{year}$) of an article is an important metric, particularly for identifying high-impact articles in a given field. It can serve as an indicator to assist researchers in gaining insights into the most current and influential research within that field.

Table 4 displays the ten most frequently cited articles. Among these, three were published in the *Journal of Machine Learning Research* with an $IF_{2022}$ of 6.0. The article that received the highest $TC_{2022}$ was authored by a team of 16 individuals from France, Japan, Germany, the USA, and the UK. The second most cited article was written by Chang and Lin [103]. These authors also hold the distinction of being the authors of the most single-country classic papers in Taiwan. Interestingly, only two inter-institutionally collaborative articles, were authored by Phillips et al. in 2006 [104] and Depristo et al. [105]. Additionally, the citation histories of the top 41 articles, each with more than 2,000 citations of $TC_{2022}$, are illustrated in Figs. 6 to 10.

**Table 4**. The top ten most frequently cited articles in ML research

| Rank ($TC_{2022}$) | Rank ($C_{2022}$) | Title | Countries | References |
|---|---|---|---|---|
| 1 (30,449) | 1 (7,859) | Scikit-learn: Machine learning in Python | France, Japan, Germany, USA, UK | Pedregosa et al. [102] |
| 2 (23,980) | 12 (1,023) | LIBSVM: A library for support vector machines | Taiwan | Chang and Lin [103] |
| 3 (19,859) | 4 (3,788) | Dropout: A simple way to prevent neural networks from overfitting | Canada | Srivastava et al. [100] |
| 4 (15,074) | 2 (4,817) | Random forests | USA | Breiman [22] |
| 5 (10,840) | 8 (1,264) | An introduction to ROC analysis | USA | Fawcett [106] |
| 6 (9,995) | 11 (1,114) | Maximum entropy modeling of species geographic distributions | USA | Phillips et al. [104] |
| 7 (9,923) | 5 (2,085) | A survey on transfer learning | China | Pan and Yang [107] |
| 8 (7,745) | 13 (1,018) | Statistical comparisons of classifiers over multiple data sets | Slovenia | Demšar [108] |
| 9 (7,672) | 97 (130) | Cognitive radio: Brain-empowered wireless communications | Canada | Haykin [109] |
| 10 (6,936) | 25 (656) | A framework for variation discovery and genotyping using next-generation DNA sequencing data | USA | Depristo et al. [105] |

$TC_{2022}$: number of citations from Web of Science Core Collection since publication year until the end of 2022.
$C_{2022}$: number of citations in 2022.

The ten most frequently cited articles are shown in Table 4. Three of the ten articles were published in the *Journal of Machine Learning Research* ($IF_{2022} = 6.0$). The most frequently cited article was published by 16 authors from France, Japan, Germany, the USA, and the UK. The second most cited article was published by Chang and Lin [103]. Lin also published most single-country classic papers in Taiwan as a corresponding author [110]. Only two articles by Phillips et al. [104] and Depristo et al. [105] were inter-institutional collaborations. The citation histories of the top 41 articles with $TC_{2022}$ more than 2,000 citations are shown in Figure 9, Figure 10, Figure 11, Figure 12, and Figure 13, respectively.

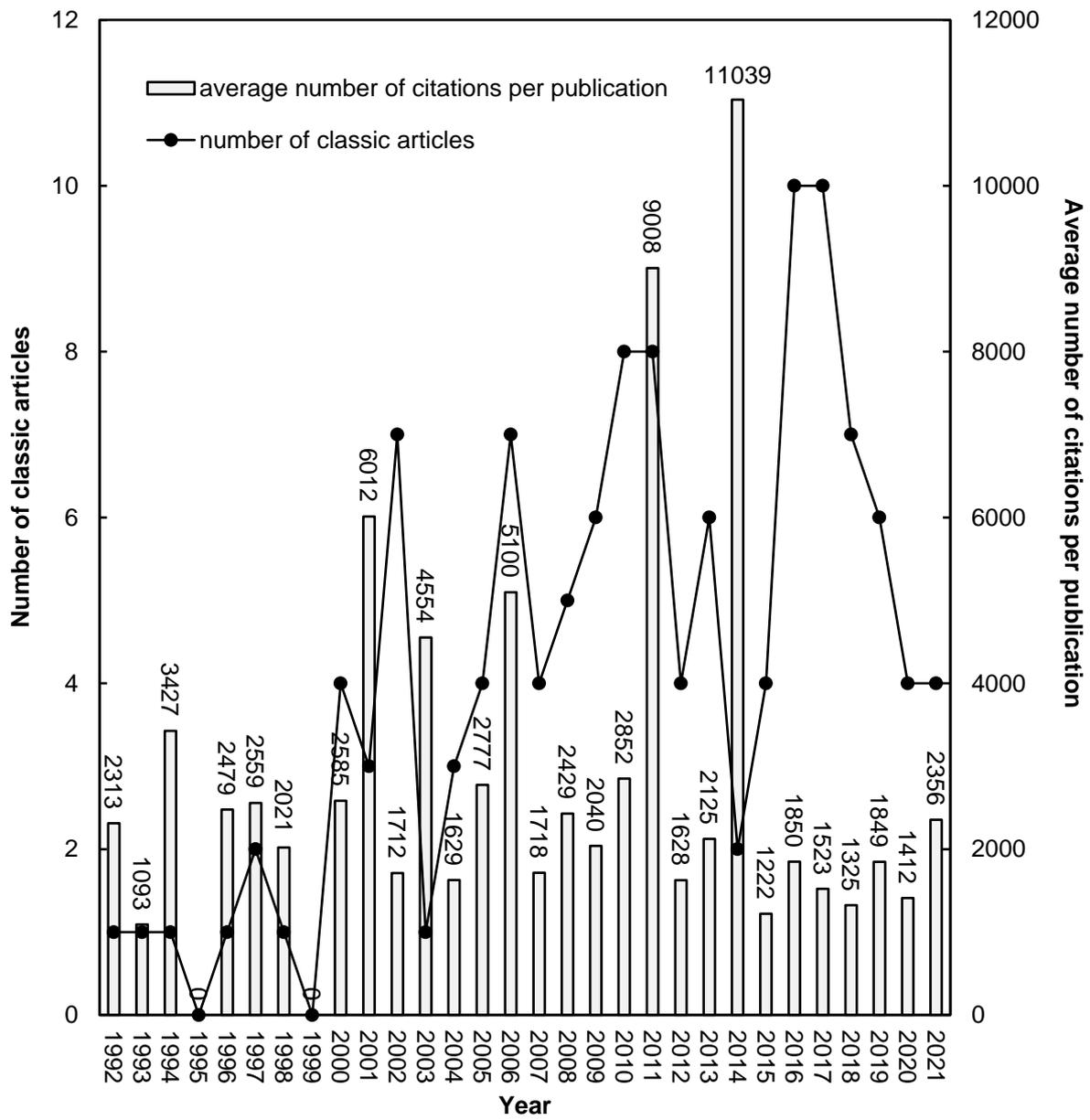

**Figure 9**: Number of classic articles and average number of citations per publication by year.

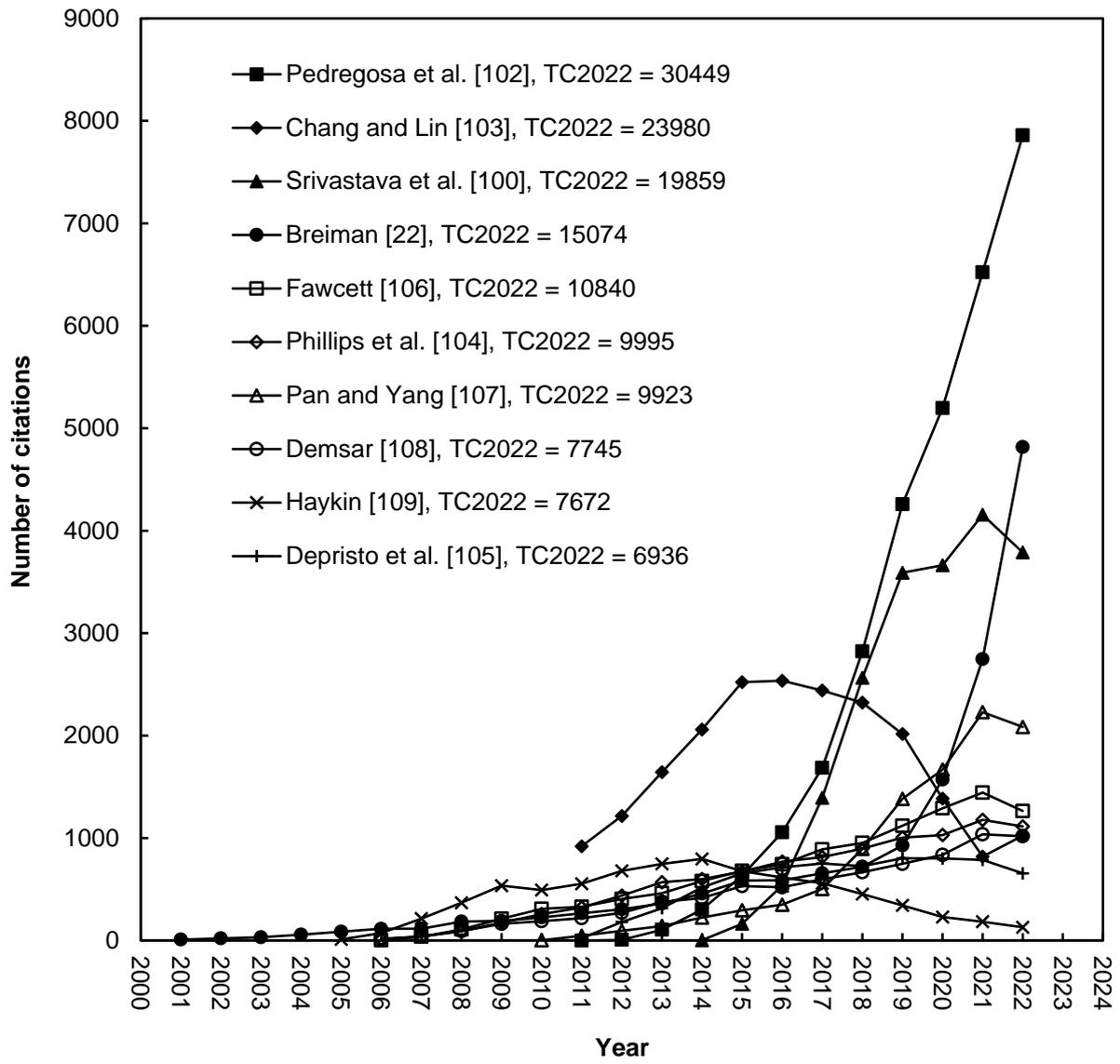

**Figure 10**. Top ten most frequently cited articles in ML research

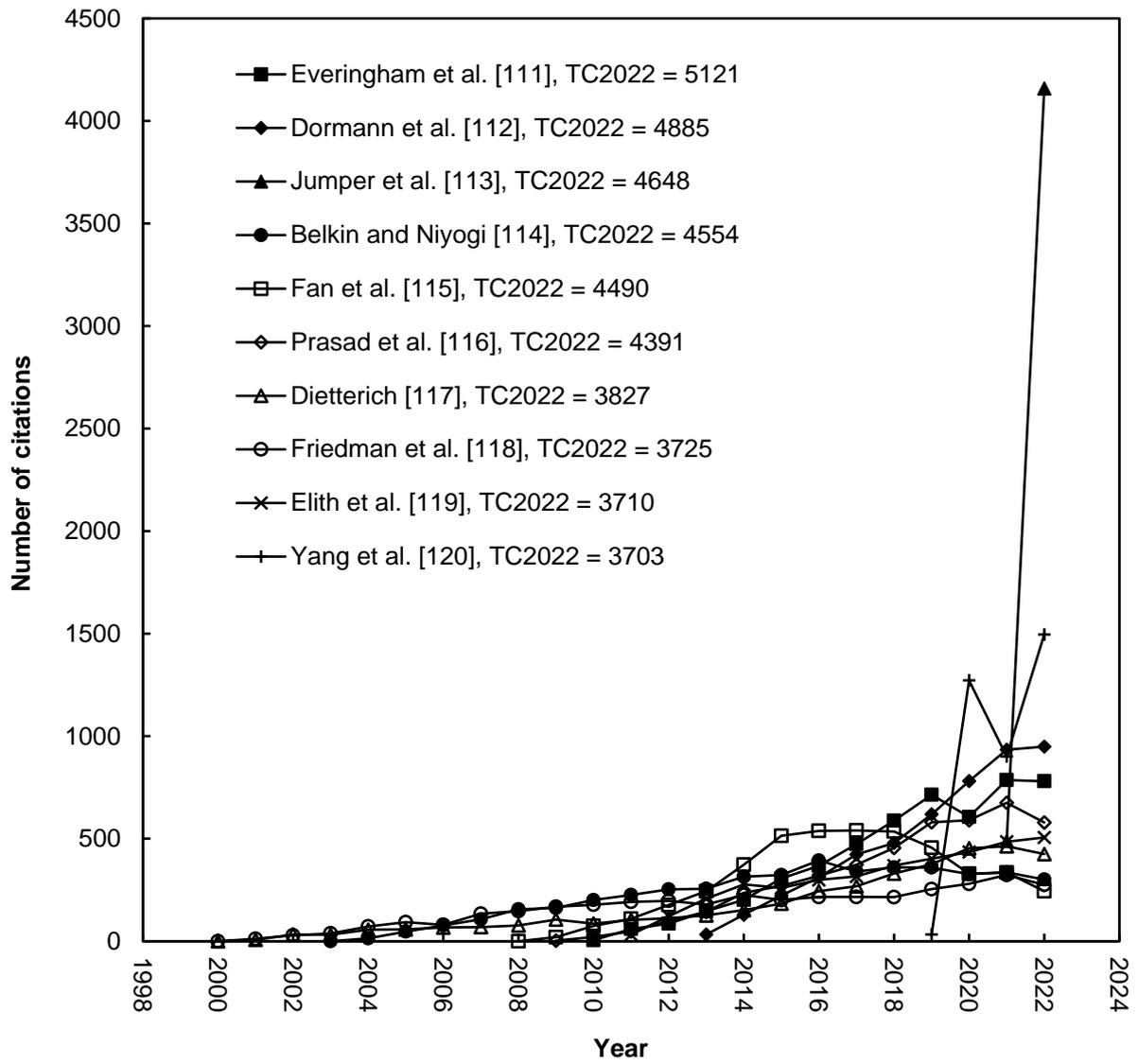

**Figure 11**. Top 11 to 20 most frequently cited articles in ML research

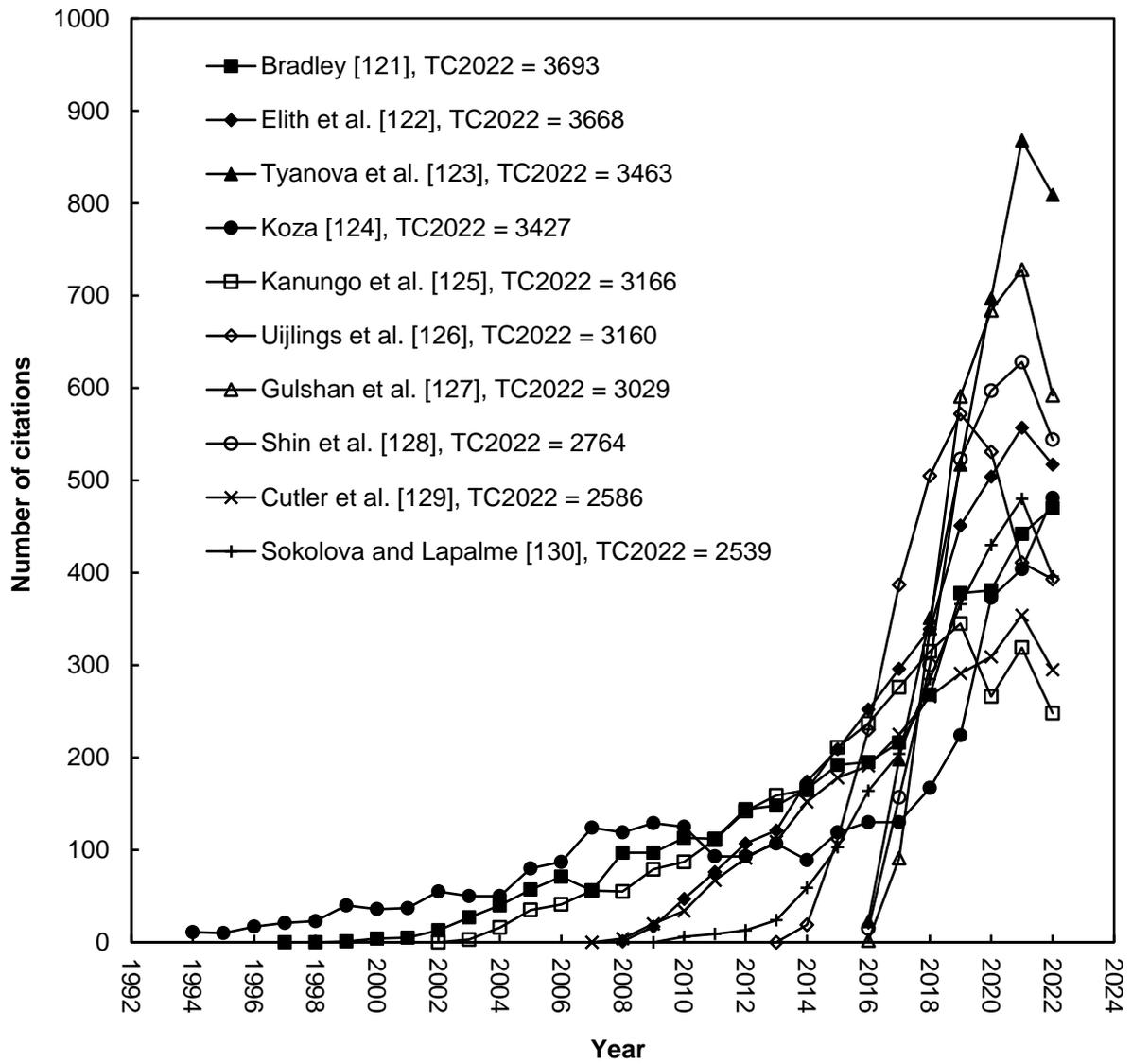

**Figure 12**. Top 21 to 30 most frequently cited articles in ML research

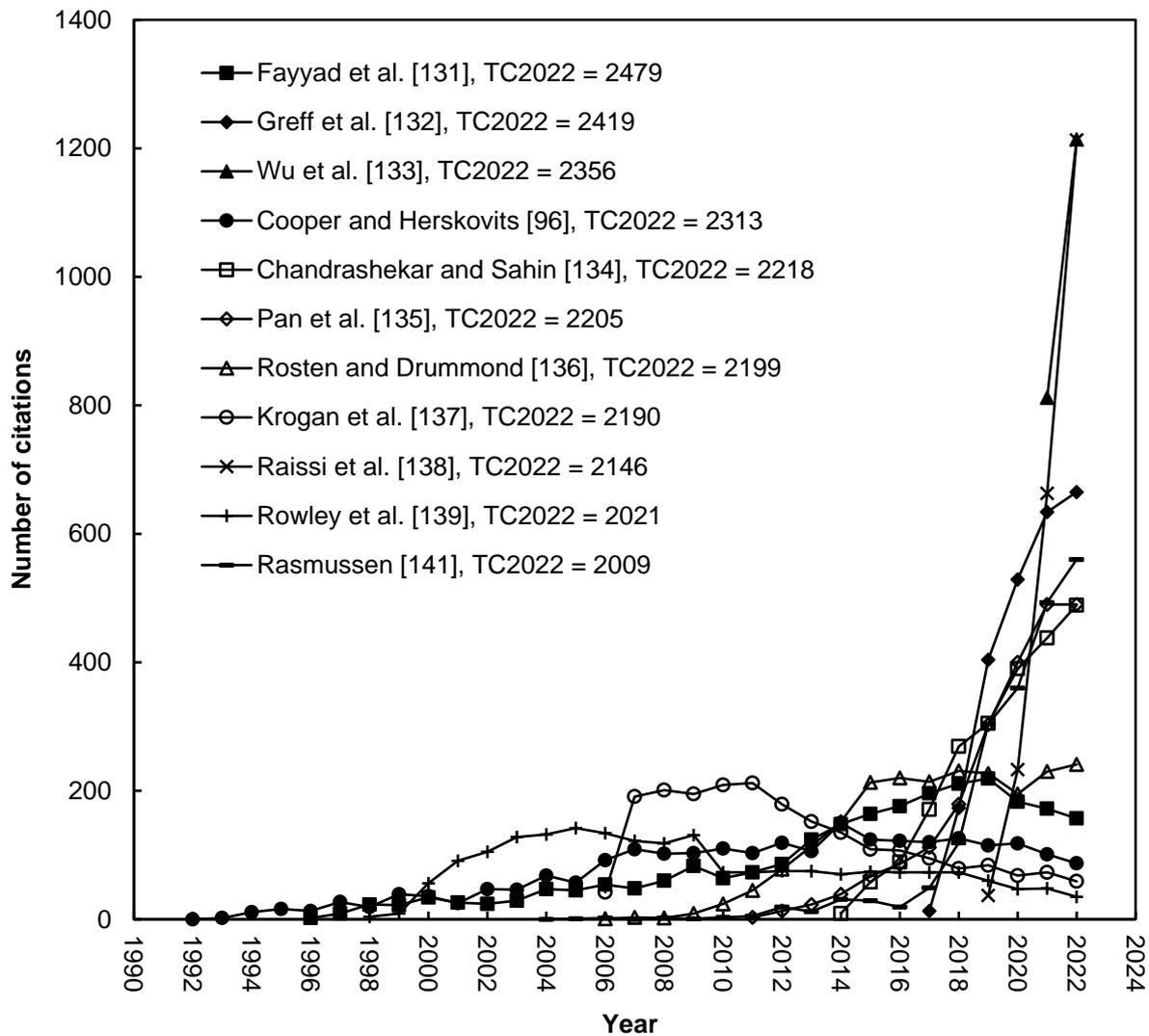

**Figure 13**. Top 11 to 41 most frequently cited articles in ML research

**4.2. Discussion**

Several articles that were illustrated in Figs. 9 to 13 hold particular significance, as they represent the most frequently cited works within their respective categories based on ML research perspectives. As such, we have provided concise summaries of select articles that we deem critical contributors to the exponential growth and advancement of research efforts in the field of ML. These summaries follow below subsequently.

Chang and Lin [103] in their study presented the implementation details of LIBSVM, a library for support vector machines, which is one of the most widely used SVM software packages. It describes the various SVM formulations supported in LIBSVM, including C-SVC, $v$-SVC, one-class SVM, and $v$-SVR, as well as the optimization algorithms and techniques used to solve the corresponding quadratic problems efficiently. Specifically, it uses a decomposition method combined with sequential minimal optimization to conquer the difficulty of large dense kernel matrices. The paper also discusses probability estimates, parameter selection tools, and other practical issues in SVM modeling. LIBSVM has been instrumental in the widespread adoption of support vector machines for classification and regression tasks, and it continues to be an important ML tool due to its high performance and ease of use.

Jumper et al. [113], in their work entitled: Highly accurate protein structure prediction with AlphaFold, published in *Nature* presents AlphaFold, a new deep learning method for predicting the three-dimensional structures of proteins from their amino acid sequences alone, with unprecedented accuracy. AlphaFold incorporates advances in neural network architecture and training procedures that integrate evolutionary and physical principles of protein structure into the model. When evaluated in the blind test of CASP14, AlphaFold produced protein structure predictions that were significantly more accurate than any previous method, with a median backbone accuracy of 0.96 Å compared to 2.8 Å for the next best approach. This represents a major breakthrough, as AlphaFold can now routinely predict protein structures at near-atomic accuracy, even without a template structure. This high-performance protein structure prediction has important applications for structural biology and drug discovery. The innovations in AlphaFold establish new benchmarks for using ML to tackle challenging scientific problems by leveraging relevant biological and physical knowledge.

Wu et al. [133], also in 2021 presented a comprehensive survey on graph neural networks, which was published in *IEEE Transactions on Neural Networks and Learning Systems*. This comprehensive survey paper provides a thorough overview of the emerging field of graph neural networks (GNNs). The paper systematically categorizes GNNs into four main types - recurrent GNNs, convolutional GNNs, graph autoencoders, and spatial-temporal GNNs - and discusses representative models within each category. It compares various GNN models in terms of architecture, input sources, pooling/readout operations, computational complexity, and applications. Significantly, the survey addresses the limitations of existing GNNs and proposes promising future research directions. This is a valuable paper as it serves as a one-stop reference guide for researchers to understand the present capabilities and future potential of GNNs, which have become increasingly important for ML tasks involving graph-structured data like social networks, knowledge graphs, and molecular datasets. The broad coverage and discussion of open challenges in the paper help outline the evolving landscape of GNN research.

While in a different study conducted by Raissi, Perdikaris, and Karniadakis [138], which is entitled Physics-informed neural networks: A deep learning framework for solving forward and inverse problems involving nonlinear partial differential equations, published in the *Journal of Computational Physics*, the authors introduce physics-informed neural networks (PINNs), a deep learning framework for solving forward and inverse problems involving nonlinear partial differential equations (PDEs). PINNs incorporate known physical laws, such as conservation principles, by differentiating the neural networks to enforce constraints based on the governing PDEs. This allows PINNs to be trained on sparse datasets while providing accurate solutions. The paper proposes continuous and discrete time PINN models and applies them to problems in fluid dynamics, quantum mechanics, and other domains. The key significance of this work is that it combines neural networks with scientific computing, leveraging deep learning capabilities for function approximation while respecting known physics. PINNs represent a novel data-efficient approach for PDE-based scientific machine-learning tasks like simulation, prediction, and system identification. They constitute an important step towards enriching modern ML with the powerful prior knowledge encapsulated by mathematical physics.

Another highly cited article includes the study presented by Greff et al. [132], entitled LSTM: A search space odyssey, published in *IEEE Transactions on Neural Networks and Learning Systems*. Their research presents a large-scale empirical study focusing on the importance of various computational components of long short-term memory (LSTM) networks. The authors compare the standard LSTM architecture against 8 variants, each differing by a single modification, on 3 benchmark tasks. They find that removing the forget gate or output activation significantly hurts performance, while coupling the input and forget gates or removing peepholes has little effect. This suggests that the forget mechanism and output nonlinearity are critical to LSTMs. Notably, the paper conducts the largest empirical study on LSTMs to date through over 5,000 experiments, demonstrating the significance of different components. The findings continue to provide useful guidance for LSTM architecture design today, as LSTMs remain a dominant model for sequence learning problems.

In the work presented by Rasmussen [140, 141] entitled Gaussian processes in ML, which has been cited more than 33396 times. This paper provides a thorough introduction to the field of pattern classification and ML. It covers many fundamental concepts, including feature selection and reduction techniques, clustering algorithms, classification methods, and evaluation criteria. Together, these components describe the typical workflow in supervised and unsupervised learning problems. The paper compares common algorithms across each component, highlighting strengths and weaknesses. While published in 2004, this review remains highly relevant as an overview of classical ML topics that still see widespread use today. Newer deep learning methods build upon but do not replace many of the foundations described here, such as pre-processing, model selection, and performance evaluation. This paper, therefore, serves as a useful introduction and reference for core ML concepts.

Lastly, Chandrashekar and Sahin [134] presented a survey study on feature selection methods, that was published in *Computers & Electrical Engineering* in 2014. The authors in their contributive efforts strived to provide a comprehensive survey of different feature selection methods for supervised learning problems. The survey study categorizes feature selection techniques into filter, wrapper, and embedded methods. It describes various ranking criteria used in filter methods like correlation, mutual information, and RELIEF. It elucidates sequential and heuristic search algorithms used in wrapper methods including SFFS, GA, and PSO. It also discusses embedded methods that perform feature selection incorporated with classifier training like SVM-RFE. The paper showcases the applicability of these methods on standard datasets. This paper provides a valuable overview of existing feature selection approaches and serves as an important reference for researchers working in ML and pattern recognition domains where dealing with high dimensional data is common. The feature selection techniques discussed remain relevant areas of active research, especially with growing data dimensions in modern data.

Although some recently published articles within the past few years had great potential, they did not rank at the top on $TC_{2022}$ but were the most impactful in the most recent year of 2022, for example, articles by Tamura et al. [142] with a $TC_{2022}$ of 1,406 citations (ranked 75$^{th}$) and $C_{2022}$ of 1,348 citations (ranked 7$^{th}$). Similarly, articles entitled "Physics-informed neural networks: A deep learning framework for solving forward and inverse problems involving nonlinear partial differential equations" by Raissi et al. [138] with a $TC_{2022}$ of 2,146 citations (ranked 39$^{th}$) and $C_{2022}$ of 1,213 citations (ranked 10$^{th}$) (Figure 12) and "A comprehensive survey on graph neural networks" by Wu et al. [133] with a $TC_{2022}$ of 2,356 citations (ranked 33$^{rd}$) and $C_{2022}$ of 1,214 citations (ranked 9$^{th}$) (Figure 12).

Classic ML articles might not have high citations in early years, for example, the article entitled "Gaussian processes in machine learning" by Rasmussen [141] with a $TC_{2022}$ of 2,009 citations (ranked 41$^{th}$) and $C_{2022}$ of 560 citations (ranked 32$^{th}$) (Figure 12). It is indeed arguable to say that classic ML articles might not have had high citations in their early years due to factors such as limited computational resources, smaller datasets, and a lack of awareness about their potential applications. However, in recent years, there has been exponential growth in their citations due to advancements in algorithms, the availability of large-scale datasets, the need for interpretable models, the rise of resource-constrained environments, and increased accessibility to educational resources. These factors have sparked renewed interest in classical ML algorithms, leading to their recognition and adoption in various domains, resulting in a surge of citations for these articles.

A concrete example of one classic ML article that was not so popular in the past few decades, is the article entitled "Neural network-based face detection" by Rowley et al. [139]. This paper presents a neural network-based system for detecting upright frontal faces in grayscale images. Multiple neural networks are trained on labeled face and non-face examples to act as filters that examine windows of the input image. A bootstrapping approach is used to iteratively select harder negative examples during training to refine the classifier boundaries. Output from the networks is merged using heuristics like thresholding and overlap elimination before arbitrating between networks. The system achieves a 90.5% detection rate on a testing set with acceptable false positives. This research shows one of the early successful applications of neural networks for computer vision tasks. While deep learning models are now more commonly used for tasks like face detection, this work formed an important foundation and demonstrated the viability of neural networks for visual pattern recognition problems.

It was found that the top ten classic articles on $TC_{year}$ and $C_{year}$ in a research topic were never the same, for example, *Helicobacter pylori* [98] with eight articles, cervical cancer with six articles, and apoptotic [99] with two articles ranked the top ten on both the $TC_{year}$ and $C_{year}$. Five articles were not only the most frequently cited with the top ten $TC_{2022}$ but also the most impactful in the recent year 2022 with $C_{2022}$ in ML research. They were summarized and further illustrated in Figure 10

*Scikit-learn: Machine learning in Python (Pedregosa et al. [102])*

The article was published in the *Journal of Machine Learning Research*, with a $TC_{2022}$ of 30,449 citations (rank 1$^{st}$) and a $C_{2022}$ of 7,859 citations (rank 1$^{st}$). This paper introduces scikit-learn, an open-source ML tool for Python. It implements a wide range of algorithms for classification, regression, clustering, dimensionality reduction, and model selection. A key goal of scikit-learn is to provide a simple and consistent interface for these algorithms while maintaining high computational efficiency. It focuses on ease of use, clear documentation, minimal dependencies, and extensive testing. The paper demonstrates that scikit-learn can achieve performance comparable to specialized ML toolboxes while providing an accessible environment for non-experts through the Python language. The scikit-learn library has since become hugely popular, with over 800 contributors supporting its continued development. It lowers the barrier to applying ML and remains one of the most widely used tools for research and applications today due to its balance of usability, extensibility, and robust implementation of cutting-edge algorithms. The article presented the initial design and implementation of this influential open-source project.

Furthermore, Pedregosa also published a collaborative article about Python entitled "SciPy 1.0: Fundamental algorithms for scientific computing in Python" [143] in the Web of Science category of biochemical research methods with a $TC_{2022}$ of 8,439 citations (and in 2024, it has recorded a total citation of 26194). This paper provides an overview of the SciPy library and its development from its initial release in 2001 to the major milestone of SciPy 1.0 in 2017. It describes SciPy's architecture and implementation choices, highlights some key technical improvements made in recent years, and discusses ongoing efforts to improve testing, benchmarking, and sustainability of the project. The paper is highly significant as it captures the growth of the widely popular and influential SciPy library, which provides fundamental algorithms and numerical routines for scientific computing in Python. SciPy is a crucial piece of infrastructure for the Python ecosystem and has had a profound impact on advances in many areas of science, being extensively relied upon in fields like data science and ML. As Python has become one of the most popular languages for research, SciPy continues to be tremendously important for powering a huge amount of cutting-edge work across domains.

*Dropout: A simple way to prevent neural networks from overfitting (Srivastava et al. [100])*

The article was published in the *Journal of Machine Learning Research*, with a $TC_{2022}$ of 19,859 citations (rank 3$^{rd}$) and a $C_{2022}$ of 3,788 citations (rank 4$^{th}$). Notable the paper has a current total citation of 50167 in 2024. This classical ML paper introduces Dropout, a simple yet effective regularization technique for deep neural networks. During training, Dropout randomly drops units (along with their connections) from the neural network to prevent units from co-adapting too much. At test time, Dropout provides a way to approximately combine exponentially many neural networks efficiently through a single network with reduced weights. The paper demonstrates that Dropout significantly improves the performance of neural networks across many benchmark datasets in domains like vision, speech recognition, document classification, and computational biology, outperforming other regularization methods. This paper is highly significant as Dropout revolutionized the training of deep neural networks and remains one of the most important techniques used today. By addressing the overfitting problem through model averaging, Dropout enabled the practical training of much larger and more complex neural networks. It has since been widely adopted and helped drive progress in deep learning, contributing to research that achieves state-of-the-art results across many domains. The simplicity yet powerful regularization effect of Dropout demonstrates how small algorithmic ideas can have a tremendous impact.

*Random forests (Breiman [22])*

The article was published in *Machine Learning*, with a $TC_{2022}$ of 15,074 citations (rank 4$^{th}$) and a $C_{2022}$ of 4,817 citations (rank 2$^{nd}$). As of 2024, this paper has recorded a total of 126363 citations. This classic ML paper introduces the method of RFs for classification and regression. RFs grow an ensemble of decision trees using randomly selected inputs at each node split to reduce the correlation between trees, thus achieving superb predictive accuracy. The paper provides a theoretical analysis showing generalization error depends on the strength of individual trees and the correlation between trees. Extensive empirical results on numerous datasets demonstrate RFs often outperform boosting while being more robust and efficient to compute. The paper is highly significant as RFs became a foundational ensemble method in ML, widely used today for problems like computer vision, natural language processing, and more. By combining randomness, decision trees, and ensemble methods, the paper laid important groundwork for state-of-the-art ML algorithms.

*An introduction to ROC analysis (Fawcett [106])*

The article was published in the *Pattern Recognition Letters*, with a $TC_{2022}$ of 10,840 citations (rank 5$^{th}$) and a $C_{2022}$ of 1,264 citations (rank 8$^{th}$). However, the paper has recorded 24505 as of 2024. This paper serves as an introduction to receiver operating characteristic (ROC) curves, which are a useful technique for visualizing and evaluating classifier performance. ROC curves allow visualization of the tradeoff between true positive and false positive rates for classification tasks. They have advantages over simple accuracy measures as they are unaffected by class imbalance or error costs. The paper discusses important concepts regarding ROC curves like the ROC space, how curves are generated from probabilistic classifiers, and how to efficiently compute ROC curves. It also highlights best practices for using ROC curves like considering the ROC convex hull to identify optimal classifiers. This paper is highly significant as it helped popularize the use of ROC curves for evaluating ML models, which are now ubiquitous in the field. ROC analysis provides a robust means of assessing classifiers that overcome the limitations of simple accuracy metrics. It allows visualization and comparison of model performance independent of dataset characteristics like class imbalance. Given the prevalence of real-world classification problems with skewed data, cost-sensitive settings, and interest in comparative model evaluation, ROC curves as introduced in this article have become an essential tool for ML research and practice.

*A survey on transfer learning (Pan and Yang [107])*

The article was published in the *Pattern Recognition Letters*, with a $TC_{2022}$ of 9,923 citations (rank 7$^{th}$) and a $C_{2022}$ of 2,085 citations (rank 5$^{th}$). This classic paper by Pan and Yang provides a comprehensive survey of transfer learning techniques developed up until 2010. Moreover, it has recorded as of 2024 a total citation count of 24165. The authors categorize transfer learning based on different settings - inductive, transductive, and unsupervised - and analyze various approaches to transfer learning including instance transfer, feature representation transfer, parameter transfer, and relational knowledge transfer. The paper provides unified definitions of key concepts in transfer learning like domains and tasks. It reviews important early works in each transfer learning setting and approach. This survey paper was highly significant as it was one of the first to systematically define the problem of transfer learning, categorize existing work, and identify open challenges. It helped establish transfer learning as an important new subfield of ML. Even today, over a decade since its publication, Pan and Yang's survey serves as a valuable reference for researchers. It provides a foundational understanding of transfer learning and highlights different problems, assumptions, and techniques. This helped spur much new research on transfer learning and its applications across multiple domains. The paper therefore plays a pivotal role in the development of transfer learning as a mainstream approach in contemporary ML.

**5. Open Research Challenges and Opportunities Relative to Global South Regions**

Data availability in quality and quantity is crucial for the performance and implementation of ML. The dominance of research output in ML in the global North is evident in this claim. Unlike the global North, many regions in the global South have yet to gain recognition in the field of ML research due to various factors such as challenges related to data acquisition and accessibility, limited computational resources, and a lack of essential knowledge of ML concepts [5, 82].

The availability of a constant electricity supply is a critical requirement for the successful implementation of ML in many regions in the global South. In contrast, stable electricity supply in global North communities has enabled the deployment of ML solutions in areas such as smart cities, smart manufacturing, and driverless vehicles. The success of ML and other artificial intelligent technologies in the global North is due to sustainable electricity infrastructure and advanced computing resources. Supercomputers, cluster computers, and internet facilities are instrumental in generating relevant domain-specific datasets for training ML models. However, the lack of these computing resources remains a significant obstacle to the widespread adoption of ML technologies in the global South.

Power supply is one of the major factors for industrial attraction. The instability in the power supply, which is common in developing regions, has not allowed the harnessing of industrial potentials in manufacturing computing devices suitable for ML applications that are peculiar to these regions. The industrial deficiencies in these regions have forced ML enthusiasts in the global South areas to rely on imported computer devices and gadgets in the development, training, and deployment of ML solutions. The cost implications of sourcing Computational resources (supercomputers, cluster computers, and internet facilities) that support the operations of ML models are high. Therefore, ML professionals in these regions are limited in exploring the technology to its full potential for the technological transformation of the regions.

Government policies on data accessibility significantly limit ML research output in the global South. Unlike the global North, where data is publicly available and easily accessible for research purposes, the global South faces restrictive government policies that impose numerous formalities for data access. These barriers discourage many scholars from engaging in ML research. Furthermore, ML algorithms are computationally intensive and require high-performance computing devices, which are not readily available in the global South. Importing such devices incurs additional costs due to government import duties. Overall, these government policies are not favorable to the progress of ML research in the global South.

Another limitation to ML research output in the global South is the inadequate collection of data and the shortage of skilled personnel. ML relies on consistent and reliable data, which are parameters to be considered by experts in the field of interest. However, the reliability and consistency of data for ML in the global South are often questionable. Additionally, there are relatively few ML researchers in the global South, and among those, many struggle to keep up with the latest trends in the field. The lack of ML experts significantly impacts research output in the global South.

The global South undoubtedly holds significant potential for novel research. Despite the identified limitations in ML output, it is important to recognize that the global South is rich with opportunities for groundbreaking ML research if it can adapt these technologies to its unique challenges. Currently, much of the ML research in the global South relies on standardized data from the global North, limiting its applicability to local environments. Therefore, the global South needs to generate representative data specific to its problems to facilitate notable and innovative machine-learning research.

To fully harness the potential for novel ML research in the global South, government policies must be favorable for data acquisition, availability, and accessibility for researchers, particularly in the field of ML. Additionally, governments in these regions should implement policies such as removing import duties on ML-related computational devices or providing government-subsidized supplies. Such policies would enhance the availability of computational resources for ML researchers, enabling them to develop cutting-edge methods tailored to the region's specific challenges.

Sensors, actuators, and various smart devices, which have become integral to modern living, offer significant potential for extensive intelligent data collection, particularly in environments where human activities may be risky. However, raw data obtained from these sources often tends to be noisy or unstructured due to the diverse nature of the devices and the varying deployment environments. Achieving uniformity in the data can be challenging. Therefore, the process of annotating raw data from these sources necessitates the expertise of human experts in ML. It is essential to develop innovative ML models capable of effectively cleaning data from noisy environments and harmonizing data from various devices to ensure accurate representation during model training.

The performance of an ML model hinges on the quality and quantity of the data used for training. Intelligent data acquisition methods, capable of capturing and reflecting the diverse characteristics of regions in the global South, are crucial for adapting ML models to address complex regional challenges. Collaboration and engagement among ML experts from both industry and academia offer an opportunity to develop standardized datasets that accurately represent the diverse nature of these regions. Creative adaptations of current ML techniques or the creation of new variations thereof are vital for cleaning raw data from these areas, resulting in intelligent datasets suitable for benchmarking. Consequently, these datasets can be leveraged to train ML models aimed at addressing prevalent issues in the regions.

Machine Learning system performance also depends on the representative data and the ML algorithm used. The better the representative features of the data coupled with efficient algorithms, the better the performance of the system. However, inappropriateness in either of the duo could affect the system severely. Therefore, the representative data that appropriately captured the global South's unique challenges would inherently require a unique approach for its processing and analysis. We propose that if the existing ML methods fail to generalize well to the representative data in the global South, the ML community must devise appropriate ML algorithms or models that are best suitable.

For instance, the detection mechanism in self-driving vehicles would need modifications or even a complete rebuild to accommodate certain features specific to developing countries before it could be effectively deployed. Otherwise, the method would fail, leading to system failure. Furthermore, advancements in Natural Language Processing have expanded to include tasks such as text-to-speech, emotion detection from text or speech, and crime and lie detection from speech. However, these achievements are primarily attributed to languages prevalent in the global North. To achieve similar accuracy or performance using languages from the global South, ML algorithms must account for the linguistic diversity in these regions. Developing such algorithms would represent a cutting-edge approach to ML techniques. In addition, while Smart Healthcare thrives in the global North due to the availability, consistency, and affordability of the encompassing technologies, the situation differs in the global South and many developing countries. Internet facilities, a crucial component of Smart Healthcare, are often expensive, inaccessible, and prone to Internet downtime. Thus, a cutting-edge ML method that can handle sparse data analysis robustly will be necessary in such regions. Similarly, the creation of Smart Agriculture and Smart Cities in most regions of the Global South will require ML models that incorporate the unique features and challenges present in these regions.

Creating standardized datasets that accurately represent the unique challenges faced by regions in the global South is key to addressing specific problems on a global scale. In addition to employing cutting-edge ML methods tailored to model the unique challenges of global South regions, it's essential to consider models that can generalize effectively across diverse datasets from various origins. The quest for cutting-edge methods capable of consistently delivering efficient results regardless of the domain or research environment is ongoing.

Different ML algorithms are suitable for different tasks with specific datasets. Some algorithms are designed for feature selection while others are designed for classification or regression tasks. More so, hybridized algorithms have been developed to harness the strengths of different ML algorithms for improved performance. These algorithms can be optimized further with relevant optimization algorithms that will serve as an efficient framework for solving problems in multi-dimensional domains within the global South landscape. The optimized hybrid algorithms can serve as cutting-edge solutions with high accuracy in handling complex data from different environments for diverse applications. Innovative hybrid algorithms combined with optimization techniques have the potential to revolutionize industries and boost performance in the regions.

Transfer learning techniques offer alternative solutions to address challenges related to computational resource constraints and data limitations. Methods such as pre-trained models, fine-tuning, domain adaptation, feature extraction, model compression, and few-shot learning are viable approaches that can help alleviate issues emanating from data and resource scarcity. Utilizing a pre-trained model or employing few-shot learning techniques can mitigate the need for a large volume of data, as the model can transfer its learned knowledge to new data regardless of its size. Similarly, resource constraints can be tackled through techniques like domain adaptation, model compression, and feature extraction. These transfer learning methods leverage existing information from pre-trained models to prevent computational complexity and optimize resource usage. A suitable transfer learning technique, carefully chosen to address data and resource deficiencies while maintaining performance accuracy, will be a cutting-edge ML approach in regions facing such challenges in the global South.

## 6. Conclusions

In this comprehensive bibliometric study, we embarked on a journey through the annals of ML's rich history to understand the foundations, evolution, and enduring relevance of classic publications. Our analysis has unveiled valuable insights that not only inform the field's historical context but also guide future research and shed light on the transformative power of classic ML literature. Our examination of the landscape of classic ML publications has led to several key findings, namely the identification of fundamental contributions that pinpointed the most influential papers and authors in the field, highlighting the seminal work that has shaped the trajectory of ML. These timeless contributions continue to serve as beacons of knowledge for domain experts and researchers in other related fields. More so, the critical analysis study presented has revealed the dynamic and evolving nature of collaborative networks within the ML community. These networks, forged through co-authorship and interdisciplinary research, have been instrumental in advancing the field and fostering innovation. Similarly, our analysis has also identified enduring research themes and emerging topics that have recently gained significant attention. This understanding of research trends provides insights into the ever-evolving nature of ML and offers guidance for future research directions. By examining the geographical distribution of highly cited publications, we have highlighted the dominance of certain countries in the realm of ML research, underscoring the global impact of this field and its ability to transcend borders.

Concerning future research directions, we believe that by building on the insights gained from this study, several future research directions that can further advance the understanding of classic ML publications and continue to shape the field's trajectory can be proposed as follows: A deeper exploration of temporal trends in classic ML publications can provide a nuanced understanding of how research themes and influential authors have evolved. This analysis can help researchers anticipate future trends and developments in the field. Investigating the cross-pollination of ideas between ML and other domains can uncover opportunities for interdisciplinary research. This can lead to the development of novel applications and solutions in emerging fields. While citation analysis is a valuable metric, future research can explore alternative measures of impact, such as real-world applications, patents, and societal impact, to provide a more holistic view of the influence of classic ML publications. As we delve into the future, promoting open access to classic ML publications and encouraging knowledge sharing can facilitate wider dissemination of critical insights, fostering collaboration and innovation. Exploring the historical context and ethical considerations surrounding classic ML publications is essential. Understanding the societal and ethical implications of foundational works can guide responsible AI development. The study of classic ML publications can inform the development of educational curricula for AI and ML, ensuring that students and future generations understand the foundational principles of the field.

**Competing interests**

The authors have no conflicts of interest, financial or otherwise, to declare.

**Data Availability**

All data generated or analyzed during this study are included in this article.

# AUTHOR CONTRIBUTIONS

**Absalom E. Ezugwu**: Conceptualization, methodology, data curation, investigation, validation, writing review and editing, project administration, resources, formal analysis, visualization, software, writing original draft, supervision.

**Yuh-Shan Ho**: Conceptualization, methodology, data curation, investigation, validation, resources, formal analysis, visualization, writing original draft.

**Ojonukpe S. Egwuche**: Writing review and editing, writing original draft.

**Olufisayo S. Ekundayo**: Writing review and editing.

**Annette Van Der Merwe**: Resources, writing review and editing, supervision.

**Apu K. Saha**: Visualization, writing review and editing.

**Jayanta Pal**: Visualization, writing review and editing.

**AUTHOR BIOGRAPHY**

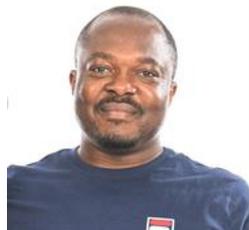

**ABSALOM EL-SHAMIR EZUGWU** completed his B.Sc. degree in mathematics with computer science, followed by the attainment of M.Sc. and Ph.D. degrees in computer science from Ahmadu Bello University in Zaria, Nigeria. Presently, he holds the position of full Professor of Computer Science within the Unit for Data Science and Computing at North-West University in Potchefstroom, South Africa. He has contributed significantly to the academic community through the publication of numerous articles in internationally refereed journals, edited books, conference proceedings, and local journals. His research focuses on artificial intelligence, swarm intelligence, and nature-inspired algorithm design, with a specific emphasis on computational intelligence and metaheuristic solutions for real-world global optimization problems. Absalom is an active member of prominent organizations such as ACM (Association for Computing Machinery), IAENG (International Association of Engineers), and ORSSA (Operations Research Society of South Africa). His dedication to advancing the field of computer science is evident in both his academic achievements and his ongoing contributions to cutting-edge research.

Email: Absalom.ezugwu@nwu.ac.za; ORCID: 0000-0002-3721-3400

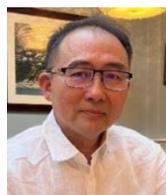

**YUH-SHAN HO**, a renowned adsorption researcher, holds a prominent position in chemical engineering with a paper amassing over 13,000 citations in the Web of Science Core Collection. He boasts 11 adsorption papers surpassing 1,000 citations each and an impressive h-index of 101 on Google Scholar. Dr. Ho's bibliometric studies have introduced six publication indicators and related citation metrics for evaluating countries and institutions. Through word cluster analysis, he uncovers research focuses and trends. Notably, he developed the Y-index, which allows for comparisons of publication potential and author characteristics. In "The World's Top 2% of Scientists" by researchers from Stanford University, Dr. Ho was ranked the top in Science Studies (Career Impact) in 2020 and 2022 worldwide, and in Information & Library Sciences (Career Impact) in 2023 worldwide.

Email: ysho@asia.edu.tw; ORCID: 0000-0002-2557-8736

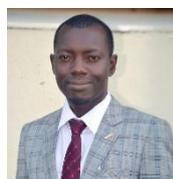

**OJONUKPE SYLVESTER EGWUCHE** obtained a B.Sc. (Hons) in Computer Science from Kogi State University, Anyigba, Nigeria in 2008. Master of Technology degree in Computer Science and a Ph.D. degree in Computer Science both obtained from the Federal University of Technology, Akure, Nigeria in 2014 and 2019, respectively. He is a Senior Lecturer at Federal Polytechnic, Ile-Oluji, Nigeria, and currently holds a postdoctoral position at North-West University, Potchefstroom Campus, South Africa. He is a registered member of the Computer Professionals Registration Council of Nigeria (CPN). His research interests include wireless sensor networks, the Internet of Things, machine learning, and AI implementations in low-computational resource environments.

Email: 55330606@mynwu.ac.za; ORCID: 0000-0002-5739-1597

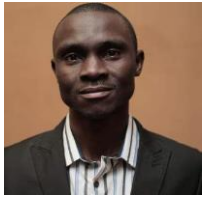

**OLUFISAYO SUNDAY EKUNDAYO** hails from Ondo State, Nigeria. He earned his B.Sc. degree in Mathematical Sciences (Computer Science option) from the University of Agriculture, Abeokuta, Nigeria, in 2008. He then obtained his M.Sc. degree in Computer Science from the University of Ibadan, Nigeria, in 2011, and later completed his Ph.D. in Computer Science at the University of KwaZulu-Natal, South Africa. In 2023, he joined the Unit of Data Science and Computing, at North-West University as a research fellow. His research interests include computer vision, affective computing, and medical imaging.

Email: 55261884@mynwu.ac.za; ORCID: 0000-0003-0203-9446

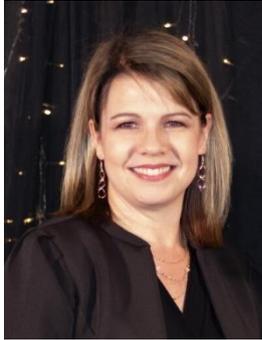

**ANNETTE VAN DER MERWE** is an established researcher in the fields of optimization techniques, artificial intelligence, and machine learning. She holds a master's degree focused on multi-objective nonlinear optimization in the field of dietetics. She has contributed significantly to the intersection of dietetics and computational methods. She received her PhD in 2019 by incorporating predictive and prescriptive learning analytics in tertiary education which addressed critical challenges in assessing and ranking student performance, offering innovative algorithms and frameworks to enhance educational evaluation systems. Dr Van der Merwe has authored numerous papers published in prestigious national and international journals and frequently presents her research on international platforms. She is recognized for her interdisciplinary expertise, bridging theoretical insights with practical solutions to complex real-world problems. Currently, Dr Van der Merwe holds the position of senior lecturer at the North-West University in South Africa, where she remains dedicated to pushing the boundaries of AI and optimization techniques, aiming to create impactful solutions that improve decision-making processes across various domains.

Email: annette.vandermerwe@nwu.ac.za; ORCID: 0000-0002-6350-6047

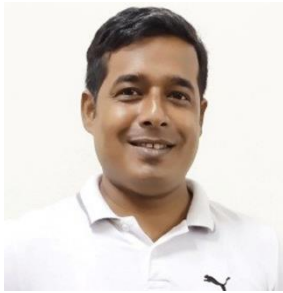

**APU KUMAR SAHA** is a full professor in the Department of Mathematics, at the National Institute of Technology, Agartala. He pursued his PhD from NIT Agartala and has been an active researcher for the past ten years. His fields of research are Artificial Intelligence, Metaheuristics, Swarm Intelligence, Multi Criteria Decision Making, etc. He has published more than a hundred and twenty articles in various reputed international journals. In addition, he published several books and book chapters and presented papers at more than fifty international conferences.

Email: apusaha_nita@yahoo.co.in; ORCID: 0000-0002-3475-018X

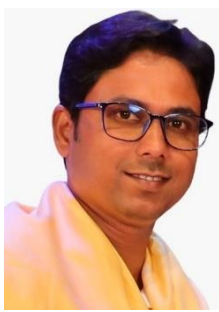

**JAYANTA PAL** has been employed as an Assistant Professor in the Department of Information Technology at Tripura University, India, since 2015. He has acquired a PhD degree from the Department of Computer Science & Engineering, NIT Agartala, Tripura, India. He completed his B. Tech. and M. Tech. in Computer Science and Engineering from IEM Kolkata, India, and Tripura University, India, in 2009 and 2011, respectively. His area of research interest includes Quantum-dot Cellular Automata, Fault-tolerant Architectures in Nano-Scale Devices, Cognitive Load Theory, and Assistive Technology.

Email: jayantapal@tripurauniv.ac.in; ORCID: 0000-0002-0719-6080